# Logic Programming on Knowledge Graph Networks
# And its Application in Medical Domain

Chuanqing Wang[1,3,4], Zhenmin Zhao[2], Shanshan Du[2], Chaoqun Fei[3,4],
Songmao Zhang[3,4], Ruqian Lu*[3,4]

University of Chinese Academy of Sciences[1], Beijing University Third Hospital[2]
CAS {Academy of Mathematics and Systems Science,
State Key Laboratory of Mathematical Sciences}[3]
Key Lab of Future Artificial Intelligence[4]
*Corresponding Author; Email: rqlu@math.ac.cn

## Abstract

The rash development of knowledge graph research has brought big driving force to its application in many areas, including the medicine and healthcare domain. However, we have found that the application of some major information processing techniques on knowledge graph still lags behind. This defect includes the failure to make sufficient use of advanced logic reasoning, advanced artificial intelligence techniques, special-purpose programming languages, modern probabilistic and statistic theories et al. on knowledge graphs development and application. In particular, the multiple knowledge graphs cooperation and competition techniques have not got enough attention from researchers. This paper develops a systematic theory, technique and application of the concept 'knowledge graph network' and its application in medical and healthcare domain. Our research covers its definition, development, reasoning, computing and application under different conditions such as unsharp, uncertain, multi-modal, vectorized, distributed, federated. Almost in each case we provide (real data) examples and experiment results. Finally, a conclusion of innovation is provided.

Keywords: Knowledge Graph Network, KGN-Prolog, Clinical Application of KGN-Prolog, Programming with KGN-Prolog on Vector Spaces, KGN-Prolog's Fuzzy, Deep and Multi-Modal Enhancement, Consultation and Argumentation as KGN-Prolog Programs, Resolution with Partial Unification.

## $ 1   Introduction

The study of medical knowledge graphs started almost at the same time when Google declared its new product: The Google Knowledge Graph [1]. The earliest paper in this aspect we found is [2] published in 2013. In this paper the authors proposed a method for constructing a knowledge graph (KG for short) of clinically related concepts. For that purpose they retrieved 95,703 de-identified EMRs collected from 17,199 different patient hospital visits of multiple hospitals. The study on this data set includes 85 topics in total. Finally, a KG called QMKG was generated automatically with 634 thousand nodes (concepts) and 13.9 billion edges (relations). Another work, the Knowlife portal published in 2014 [3], is a large KG for health and life sciences, published in 2014, automatically constructed from Web sources, including concepts of diseases, symptoms, causes, risk factors, drugs, side effects, etc. Its difference from earlier works was the technique of not only collecting patients' data from EMRs, but also from medical literature including 593,423 abstracts of Pubmed Medline and Pubmed Central, and 28,650 Web pages Web publications. This means that new KG construction does not only rely on data processing, but also involve NL processing techniques. It is of interest to mention at this place that KGs of traditional

Chinese medicine have been still a research topic nowadays, for example [4].

As for recent publications, we mention [5] which reported an approach building MKG from EMR end-to-end. Its original data set contains more than 16 million de-identified clinical visit data of more than 3.75 million patients. M Rotmensch et al. proposed an approach to generate a KG from EMR data by mapping diseases to their possible symptoms automatically [6]. Their data source includes 156 diseases and 491 symptoms from more than 273 thousand patient visits to the emergency department. Other examples include [7-10].

However, we have found that there are four main contrasts regarding AI-related medical KG development and application: Abundance of medical KG development effort vs. shortage in reports about practical achievements of medical KG application. Abundance of available general-purpose programming languages vs. shortage in special-purpose medical programming languages. Abundance of using logical techniques to do reasoning on KGs vs. deficiency of published examples of using the most popular logic programming language Prolog on medical KGs. Abundance of available independent medical KGs vs. shortage of multiple KGs group cooperating in medicine (or even in general) application. It was just these four main contrasts in medical KG development and application which has driven us to study and work on the subject denoted by the title of this article.

## § 2   Logic Programming on a Knowledge Graph Network

**2.1**   KGN-Prolog: A programming language in logic on knowledge graphs

Traditionally, a Prolog program runs on a simple theory consisting of a set of rules and predicates, where the latter constitutes the data part of the program. A more advanced form of reasoning is to let Prolog programs running on a data base, usually on a relational database. This combination of Prolog + data base is usually called deductive database. In this paper we will study a novel form of knowledge reasoning: the 'Prolog + knowledge graph' paradigm, or simply deductive KG. Based on this idea, we do a step further and introduce an even more advanced form of knowledge based logical reasoning: the Prolog + knowledge graph network (KGN) paradigm, or simply deductive knowledge graph network (DKGN). The details will be given in the next subsections.

In the case of healthcare domain, such as a hospital, the KGN may consist of four kinds of KGs: those of clinical medicine (including KGs of different clinical domains), those of hospital's medical resources (including medical test records of different modalities), those of doctors (different knowledge background, work experience and points of view) and different patients KGs (including KGs for lifelong health records of single patients and KGs for statistical information of a big amount of patients or diseases).  See Figure 2-1 for a demonstration, which displays a KGN consisting of seven KGs, where the arrows show the 'calling' direction between KGs.

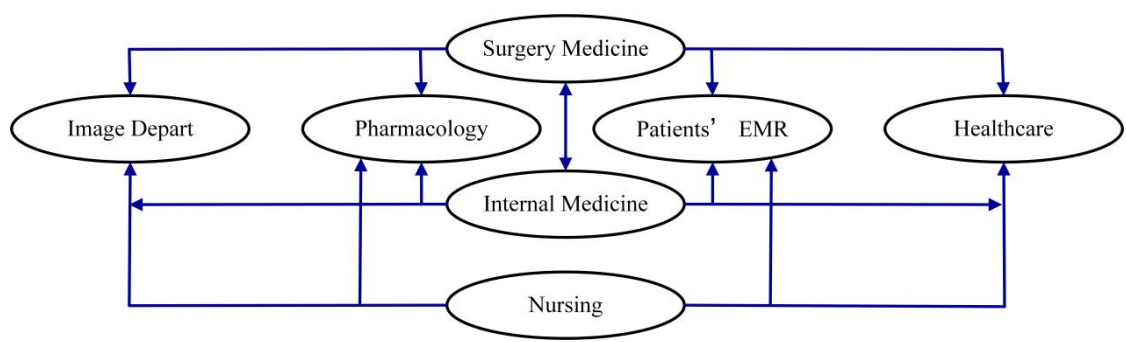

Figure 2-1 A Knowledge Graph Network

However, the design of a KGN should be done carefully. Dividing different knowledge branches into different KGs should not be the sole principle of KGN design. Sometimes, separate knowledge of different diseases may lead to breaking the natural connection between them. For example, some diseases are symbiotic. Some others have causal relation. Yet some others display similar symptoms. To separate them in different KGs reluctantly would produce negative effects.

Historically, the KGN concept and research was a proposal submitted by the authors to the Chinese ministry of science and technology as a key project [11] for the research program "Science and Technology Innovation 2030-New Generation Artificial Intelligence" lunched in 2020. Based on that proposal, this paper introduces the following new ideas and functions: (1) Decide to use Prolog language and logic programming to control a KGN. (2) Develop various new functions for KGN-Prolog to meet different application needs. (3) KGN-Prolog is decentralized such that each KG of it is attached with an independent local Prolog program (LPP). (4) This makes KGN-Prolog a deductive decentralized knowledge system. For more details see Definition 2.2 and following sections.

In the next sub-sections we will introduce KGN-Prolog Programming stepwise. First we discuss Prolog programming on a single KG. Then we extend this technique to Prolog programming on multiple KGs. Thirdly, we introduce KGN-programming on a knowledge graph network. Finally, in the conclusion section, we introduce and discuss an advanced form of KGN-Prolog programming — the federated KGN-Prolog programming.

**2.2   Basic Programming devices of KGN-Prolog Programming**

The new language of Prolog style introduced in this paper is called KGN-Prolog because its programs are running on a knowledge graph network (KGN). Knowledge graph (KG) has become a very popular knowledge representation form since its birth and has been also widely applied in many areas including the healthcare domain. However, the service capability provided by a single KG is often limited. For a huge size KG, the complexity of its computation would be very high. In addition, the consistency of its knowledge is often not guaranteed. Therefore, it is often a good choice of letting multiple KGs working together or decomposing a big KG in several cooperating sub-KGs. Besides, different domain experts in the same application domain often have different opinions towards many problems, based on their different knowledge and experience background. This makes a KG representing their opinions inconsistent. Furthermore, it is often not appropriate to put different knowledge representations in the same KG, for example knowledge in text or image form. In this paper we introduce the KGN concept to meet the need of admitting heterogeneous and/or diversing knowledge coexisting in a global knowledge cooperation and competition framework.

According to the various needs of reasoning on multiple knowledge graphs, in particular the reasoning needs on medical problems, the classical Prolog functions are extended to the KGN framework to support collective reasoning facilities on KGs, which will be introduced one by one in the following of this paper. As a first step, we will give a preliminary description of how our KGN-Prolog works on a single KG.

**Definition 2.1**: We first introduce additional functions and predicates to classical Prolog for reasoning on a single KG. They include, but are not limited to, the following ones:

**Functions**:

RDF (X, Y, Z): It denotes the RDF triplet consisting of an ordered 3-sequence X, Y, Z.

Head (W), Relation (W), Tail (W) are three functions denoting the head, relation, respectively tail term of the RDF triplet W.

**Predicates**:

Tr1 (W): It means W is a RDF triplet.

Tr2 (X, Y, Z): It means (X, Y, Z) is a RDF triplet.

Y (X, Z): Only for comparative relations Y (see definition 4.3). It means Y is the relation of a triplet where X is the head entity and Y the tail entity.

Is-class (X, Y): X is a class or subclass of the KG Y.

Is-schema (X, Y): X is a schema or sub-schema of the KG Y.

In-class (X, Y): X is a member of the class or subclass Y.

Is-class-of (X, Y, Z): X is a member of the class Y of the KG Z.

**Rules**:

Use the notation #<KG name># before a rule or part of it to denote from which KG the data come.

**Example 2.1:** We may want to have a Prolog program which produces a snapshot for any person X, where snapshot (X) consists of all RDF triplets with the entity X as the head term and all those with X as the tail term. Given a KG containing information about some person Li, according to the conventional Prolog programming style, one can use the following rule to produce the snapshot of Li, where 'Fail' is the repeating operator:

? #<KG name># Snapshot (Li)

Snapshot (X) ← in-class (X, Person), Output (RDF (X, Y, Z)), Fail;

in-class (X, Person), Output (RDF (U, V, X)), Fail.                (2-1)

For a practical example of (2-1) on multiple KGs see the next sub-section.

## 2.3. Logic Programming on multiple KGs.

**Example 2.2:** We extend example 2.1 to the case of multiple KGs by summarizing a snapshot of the well-known film director Zhang Yimou's data from two large KGs DBpedia [12] and Yago3 [13].

Now the Prolog program looks like follows:

Query: ? #(DBpedia, Yago3) # Snapshot (Zhang Yimou).

Rule: Snapshot (X) ← Generate1 (X); Genarate2 (X).

Generate1(X) ← Output ((RDF (X, Y, Z))), Fail.                      (2-2)

Generate2(X) ← Output ((RDF (X, Y, Z))), Fail.                      □

**Experiment 2.1** For implementing the above example, we only extracted three domains of knowledge from the two KGs, which are related to Chinese persons, cities and films. We have got in total 303,046 resp. 422,132 triplets from [12] and [13]. Among them 115 triplets from [12] and 100 ones from [13] match the requirement of (2-2). Figure 2-2 shows a reduced snapshot of Zhang_Yimou composed from 15 triplets. This experiment was run on a computer with i9-13900K CPU and 64 G memory five times. The average compile time (separately done) used was 485.3 seconds and average execution time 11.8 seconds.

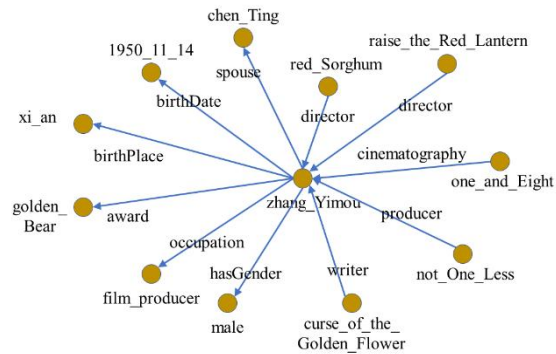

**Figure** 2-2    A reduced snapshot of Zhang Yimou

### 2.4  Logic Programming on a KGN.

**Definition 2.2:** 1. A KGN is a finite non-empty network S of KGs, each with a different name. There are four possible cases of connection between any two KGs K1 and K2 in this network. They may be a link from K1 to K2, or a link from K2 to K1, or both of K1 to K2 and K2 to K1, or no link between them at all (see Figure 2-1 for an example). Each link means a 'call' relation. For example, K1 calling K2 means K1 acquires knowledge from K2.  KGN-Prolog program running on S consists of a finite non-empty set of local Prolog programs (LPPs), where each of them is attached to one KG (Thus for the same KG, the KG body and its LPP are each other's partner) with the limitation that each KG may at most have one partner LPP. The LPP of a KG may have two data parts. The first one is its own data as part of a Prolog program. The second one is the set of all triplets in its partner KG. Each time during reasoning, the LPP first checks its own data part, and then its partner KG's data part. Besides these, it may also 'call' another KG and/or that KG's partner LPP for extra rules and data. The above mentioned links between KGs in a KGN denote their calling (visiting) relations.

**Example 2.3:** The medical community has a general recognition that those people will very likely get cardiovascular disease if they have unhealthy habits, increased blood pressure and atherosclerotic plaque, etc. where unhealthy habits mean people with age exceeding 60 like smoking and drinking, having bad BMI and less physical activity [14-16].

We consider the data set [17] which collects more than 700,000 person's data regarding their possibility of suffering from cardiovascular diseases, where each person's data include 11 triplets as attributes characterizing this person's healthy. They are age, height, weight, gender, systolic blood pressure, diastolic blood pressure, cholesterol, glucose, smoke, alcohol-intake and physical activity. For example (68001, weight, 60 kg) is a triplet, where 68001 is the person number. Since the three attributes — height, weight, and gender — cannot individually explain a person's constitution by themselves, it is more rational to use a new attribute bad-BMI to replace them three, where BMI=height/(weight*weight) was introduced by [18].

**Experiment 2.2**: We want to know which ones from the remaining 9 triplets influence human's healthy at most. In the following we define an Atriplet as the 9 attribute triplets plus one label attribute 1 (0) to mean a positive (negative) example of cardiovascular disease (whether or not suffering from the disease). More exactly, we want to know the order of seriousness (curse-rate) of bad impacts of these different attributes.

The program is the following, where we use a new facility of KGN-Prolog programming introduced by us: the repeating operator Fail may have a parameter as index showing the times of loop this rule may at most perform (Otherwise the fail operator may run an infinite number of times if not affected by other factors). For example, Fail (9) means it may repeat at most 9 times. The second new facility is the expansion of data definition. In order to avoid accessing a long list of data predicates within a loop, we allow a large set of data elements used in functions or predicates to be listed in Data part of a KGN-Prolog program

**Rule:**

Job Done ← Get (Buffer, I), Process (Buffer, I).                                                (2-3)

Get (Buffer, I) ← Input (Atriplet), Positive (Atriplet), Count (I),

   Transform (Atriplet, Apair), Distribute (Apair, Buffer), Fail (5000).         (2-4)

Process (Buffer, I) ← Filter (Buffer), Display (Buffer, I, 1).                            (2-5)

Display (Buffer, I, J) ← Print (Attribute Name (J)), Print (Quotient (Volume (Buffer (J)), I),

   Increase (J), Fail (9).                                                                              (2-6)

**Data:**

**Attribute:** age, systolic blood pressure, diastolic blood pressure, cholesterol, glucose, smoke, alcohol-intake, physical activity, bad_bmi.

The program runs as follows. Rule (2-4) fetches 5000 persons' data from the KG [17] randomly, where each data is an Atriplet. After any Atriplet is introduced, it checks the Atriplet's legality. Only those data of cardiovascular patients (positive examples) will be accepted and counted by the predicate counter (I). It then transforms Atriplet's each triplet of form (attribute (person number), eq, value) into a pair (attribute, value). The inputted Atriplets are thus transformed into Apairs. There are nine buffers. The Distribute predicate dissolves each Apair in 9 pairs and distributes them in 9 different buffers respectively for doing statistics.

Since usually not all attributes in an Apair have bad values, the Filter predicate of rule (2-5) removes from each Buffer (J), 1<=J<=9, all pairs which do not have bad attribute values. The Display predicate calculates the quotients (number of pairs remaining in each Buffer / quantity I of positive examples) and outputs them.

We define the bad values of attributes as: age > 60, BMI-index difference > 25, systolic pressure > 130, diastolic pressure > 80, cholesterol not equals to 1, glucose not equal to 1, smoke, drink and less physical activity.

From the 5000 persons' data, there are 2498 cardiovascular patients (2502 not patients). During the experiment, the program took 316.75 seconds, where "get (Buffer, I)" took 316.73 seconds and "process (Buffer, I)" took 0.2 seconds. The ordering of all 9 bad attribute values (their rates over the total bad data quantity) is as follows:

systolic: 0.6988, cholesterol: 0.4692, diastolic: 0.4436, glucose: 0.3372, bad BMI: 0.219, age: 0.1968, smoke: 0.177, alcohol: 0.084, physical: 0.0524.

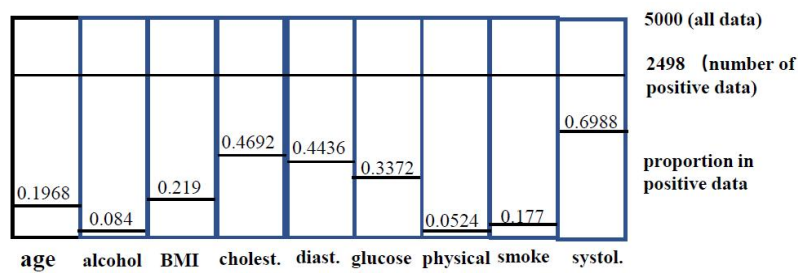

Figure 2-3    The distribution of bad attribute values

## § 3 Logic programming on KGN embedded vector spaces

### 3.1 Basics of Prolog Programming on Vector Spaces

Embedding a KG into vector space has become a very popular technique of KG computing. We extend KGN-Prolog's power for performing logic programing also on KGN's vector space representation. There are three possibilities: 1. All KGs of the KGN embedded in the same vector space. 2. Different KGs embedded in different vector spaces. 3. The combination of 1. and 2.

**Definition 3.1:** The KG-embedded vector spaces have the following major properties:

** Each vector space transformed from a KG A is named A-V with a positive integer as its dimension, where A is the name of KG;

** Each vector space is either dynamically constructed during Prolog program embedding, or uploaded from a database.

**Definition 3.2:** For doing reasoning on vector space representation of KGs, KGN-Prolog's power includes (but is not limited to) the following additional facilities:

**Declaration (optional)**: EMBA: <embedding algorithm name> {<Prolog program part>}: This declares which embedding algorithm is used in this program part.

**Built-in Functions**: VEC (X, Y, Z): It denotes the vector form of an RDF triplet consisting of the ordered 3-sequence X, Y, Z.

V-Head (W), V-Relation (W), V-Tail (W): where W is a vector, are three functions whose values are three vectors pointing to the start and end of W, respectively.

V-distance (X, Y): distance between the two vectors X and Y. Its definition depends on the structure of the vector space. For a conventional Euclid space it is defined as the value of $((x_1^2-y_1^2) + \ldots + (x_n^2 - y_n^2))^{1/2}$, given the vector space is n-dimensional, where $(x_1,\ldots x_n)$ and $(y_1,\ldots, y_n)$ are their coordinates, resp.

**Built-in Predicates**: Embed (X, Y, Y-V, n): Use Algorithm X to embed KG Y in vector form Y-V of dimension n.

V-Tr1 (W): It means W is an RDF triplet in vector form.

V-Tr2 (X, Y, Z): It means (X, Y, Z) is an RDF triplet in vector form.

# V-Y (X, Z): This is the predicate form of a triplet (X, Y, Z), where Y is a comparative relation (see definition 4.3).

KG-to-VS:

T-to-V (X, Y, Z, T): It transforms data unit X in vector form Y of the vector space Z with embedding algorithm T, where X may be in numerical, or textual, or image form, Z and T may be empty if they are unique.

V-to-T (Y, X, Z, $T^{-1}$): It transforms vector Y of vector space Z back to data unit X with reverse embedding $T^{-1}$, where Z and $T^{-1}$ are optional if they are unique.

V-Op (X, Y, Z): Z is the result of the vector operation V-Op on X and Y.

Output-V(X1,…,Xn): All Xi are vectors.

**Example 3.1:**

Reconsider example 2.1 of section 2. Assume we are mining a snapshot not from a normal KG, but from KG-V, which is a vectorized KG, where each triplet is a vector. Then the rule would be as follows:

snapshot(X) ← in-class (X, person), #KG-V# output (Vec (X, Y1, Z1)), Fail;
        in-class (X, person), #KG-V# output (Vec (X2, Y2, X)), Fail.     (3-1)

Or also in the form:

snapshot(X) ← #KG-V# (in-class (V-head (X), person)), output (V-tr1 (X)), Fail;
      in-class (V-tail (X), person), output-V (V-tr1)), Fail.   (3-2)

### 3.2 Embedding high dimensional KG in Vector Space

Usually a KG consists of a large set of knowledge triplets. But there are situations where the unit of knowledge is not a single triplet, but an ordered set of triplets. We have already met such situation in example 2.3 and experiment 2.2, where a knowledge element consists of 11 triples plus a label. This sub-section discusses how to embed such KGs in vector form and how to do knowledge completion on such vector KGs.

**Example 3.2**. Reconsider the data set [17] used in example 2.3. This time we are interested in two tasks. The first one is how to embed this kind of data sets in a vector space. The second one is, given a new patient's personal data which is of the same sort as the above-mentioned data sets, how to use this vector form KG to decide whether the new patient has got cardiovascular disease or not.

For the first task, we do the following: (1) Transform the data set into a KG, where KG's each data element is a finite sequence of m triples. m is a fixed positive integer. (2) Extend the TransE algorithm to a more powerful one TransMETH, where M means multiplicity, T means triplets, H means hash, such that for each data element, the m triples are embedded to m different local vectors with different weights in m different subspaces, whereby the embedding relation is recorded in a hash table HH. (3) The m local vectors of each data set form a global vector, to which a positive/negative label is attached.

For the second task, we proceed as follows. (1) Given any new patient X's data D, transform D in a data element N of m triples. (2) Compare N with the KG's data stored in hash table HH (triple by triple). Given a positive integer J, for each $i$, find the set S ($i$, J) of J nearest i-th-triple sets in HH and their corresponding local vectors. (3) Form a virtual global vector D-V of data D based on these local vectors. (4) Given a positive integer K, find the set S of K-nearest vectors of D-V. (5) Calculate the rate of the number of vectors in S with a positive label over the volume |S|. This is the chance that patient X suffers from cardiovascular disease. (6) Varies the value of K to get different chances with different levels of K nearest vectors. This task is completed by algorithm 3.2 of next sub-section.

Following is the TransMETH algorithm which is an enhanced form of TransE algorithm for multiple triplet sequence embedding with a hash table.

---

**Algorithm 3.1** TransMETH

**input** A set S={ $(t_1, t_2, ..., t_m, l)$ }, weights set W={$w_1, w_2, ..., w_m$}, where each m+1 tuple $(t_1, t_2, ..., t_m, l)$ is stored as an item in a hash table HH. The $w_i$, where all $w_i > 0$, $1 \leq i \leq m$, $\Sigma_i w_i = 1$, are the weights of $t_i$ respectively. Let L be the labels set, and $T_i = \{t | t$ is a triplet of i-th kind$\}$, i=1,2,...,m. $n$ is the embedding dimension of $t_1, t_2, ..., t_m$ and $l$. $\gamma$ is the margin.

1: **initialize**   $l \leftarrow$ uniform$(-\frac{6}{\sqrt{n}}, \frac{6}{n})$ for $l \in L$

2:     $t_i \leftarrow$ uniform$(-\frac{6}{\sqrt{n}}, \frac{6}{\sqrt{n}})$ for $t_i \in T_i$

3: **loop**

4:      $t_i \leftarrow t_i / \|t_i\|$ for each triplet $t_i \in T_i$
5:      $l \leftarrow l / \|l\|$ for each label $l \in L$
6:      $S_{batch}1 \leftarrow$ sample $(S(p), b1)$ // sample a minibatch of size b1 from positive examples of S

   $S_{batch}2 \leftarrow$ sample $(S(n), b2)$ // sample a minibatch of size b2 from negative examples of S, Where $S(p) \cup S(n) = S$, b1 = c $|S(p)|$, b2 = c $|S(n)|$, 0 < c <1.

7:      $Q_{batch} \leftarrow \emptyset$ // initialize the set of pairs of m-triples.
8:      for $(t_1, t_2, ..., t_m, l) \in S_{batch}1 \cup S_{batch}2$ do
9:          $(t_1', t_2', ..., t_m', l') \leftarrow$ corrupt(j) // Replace j $t_i s$ with j $t_i' s$ and $l$ with $l'$ for each m+1 tuple in $S_{batch}1 \cup S_{batch}2$, where j is randomly selected each time. $l'$ is the opposite of $l$.
10:     $Q_{batch} \leftarrow Q_{batch} \cup \{((t_1, t_2, ..., t_m, l), (t_1', t_2', ..., t_m', l'))\}$
11:     End for
12.     Update embeddings w.r.t.

$$\nabla [\gamma + d(w_1 t_1 + w_2 t_2 + ... + w_m t_m - l) - d(w_1 t_1' + w_2 t_2' + ... + w_m t_m' - l')]_+$$

13.     Store all regulated addresses of combined m+1-vectors in the HH $((w_1 t_1, w_2 t_2, ..., w_m t_m, l))$ address of the same hash table, corresponding to their original inputs.
14: **end loop**

**Experiment 3.1** Given the data set [17], we transform each data unit in form of a sequence of 11 triplets plus a label (yes or no). Each of them looks like ((68001,age,63), (68001,gender,1), (68001,height,1.54), (68001,weight,78.0),(68001,systolic blood pressure,140), (68001,diastolic blood pressure,90), (68001, cholesterol,1), (68001, glucose,1), (68001, smoke,0), (68001,alcohol intake,0),(68001, physical activity,1),1), where 68001 is the person number, the last digit 1 (0) means positive (negative). We call this form of data unit a 11-let and gave the name Cardi to this new KG. To embed Cardi in vector form Cardi-V we modified the open source TransE code of [19] to implement the idea of our TransMETH algorithm, where we set the weights of 11 attributes as 4:4:2:4:8:8:7:7:2:2:2. The dimension of vector space is 64. Learning rate = 0.001. margin = 1.0. The norm L1 was used to calculate the distance between two vectors. The sampling coefficient c = 0.04 was used for both positive and negative person data.

We use algorithm 3.1 to embed 68000 data items of Cardi to Cardi-V while leaving the remaining 2000 data items for testing the result in sub-section 3.4.

**3.3    Programming on the Vector Space**

Our experiment is done in four steps. First step: embed the KG Cardi in vector space Cardi-V, which we have done in last sub-section. Second step: Given a new patient X's data D, construct its vector form R in Cardi-V. This will be done next by algorithm 3.2. Third step: Use k nearest neighbor method with different k-values to calculate the chance that X (R) suffers from cardiovascular disease. Fourth step: Compare our approach with a statistical approach to estimate its preciseness and efficiency.

Algorithm 3.2 first excludes abnormal input data if any which are outside of the standard interval. It then builds a virtual vector R of Data (X) by comparing Data (X)'s basic information (the first 11 triplets of Data (X)) with the information stored in the hash table HH during algorithm

3.1's running (We call R as a virtual vector because it hasn't been inserted in Cardi-V by an embedding algorithm). During Algorithm 3.2's running, the J nearest HH data most close to Data (X) for some positive integer J will be used to construct a virtual vector R. This process is done by the following algorithm:

**Algorithm 3.2**    Construct a Virtual Vector R

Given the vector space Cardi-V, the attached hash table HH, a new 12-let, simply called L12, where the label information (the $12^{th}$ triplet) is empty, and an integer J > 0. Construct a global vector R for L12 in Cardi-V as follows:

**Let** L12-i to denote the i-th triplet of L12.

**For** i = 1 **to** 11 **do**

    **Begin**

    **If** the tail term t of L12-i is abnormal (See Note 3.1 below) **then**

        halt the program running and issue a warning message.

    **If** the value domain of the i-th attribute is infinite **then**

    **Begin**

        Consider the set H (J, i) of J those i-th triplets in HH, such that they are most close to L12-i;

        Fetch from HH the set HV (J, i) of J local vectors corresponding to H (J, i);

        Construct the i-th synthesized local vector V(i) as an average of HV (J, i);

    **End;**

    **Otherwise**

    **Begin**

        **If** HH contains an i-th item x with the same attribute value **then**

            let it be the attribute value of x

        **Otherwise** give up. The algorithm fails.

    **End;**

Construct the weighted sum of all local vectors V(i) constructed in the above loop to build a virtual vector R.

**End of algorithm.**

**Note 3.1:** The normal intervals of the attributes' values used in this experiment are: age/years：[0-130], gender/{masculine, female}：{1,2}, height/meter: [0.50-2.30], weight/kg：[3.0-200.0], systolic blood pressure: [60-230], diastolic blood pressure：[40-220], cholesterol level: {1,2,3}, glucose level: {1,2,3}, smoke: {0,1}, alcohol intake：{0,1}, physical activity：{0,1}.

**Example 3.3:** Represent new patient's multiple triplet data as a virtual vector in Cardi-V by algorithm 3.2 and determine its label by using k-nearest neighbor vectors. The value J is used in algorithm 3.2 for J-nearest vectors approximation.

**Rule:** cardiovascular (X, Cardi-V, K, J, Chance) ← Construct (Data (X), Cardi-V, J, R), Nearest (R, Cardi-V, K, S), Positive (S, S1), Divide (Size (S1), K, Chance). (3-3)

Perform the following Query:

**Query:** ? cardiovascular (Qian, Cardi-V, 5, 10, Chance)

**Data:** Qian = ((Qian,age,72),(Qian, gender,1),( Qian,height,1.71), ( Qian,weight,70.0), ( Qian, systolic blood pressure,138),( Qian, diastolic blood pressure,110), ( Qian, cholesterol,1), ( Qian,

glucose,1), ( Qian, smoke, 0),(Qian, alcohol intake,0), ( Qian, physical activity,1),Y).

The results are provided in next sub-section.

When running the query above, rule (3-3) will be initiated. The predicate Construct transforms patient Qian's data (with blank label Y) to a vector R in the same vector space Cardi-V by using J (already existing) nearest vectors to approximate R. The predicate Nearest performs algorithm 3.2 for J = 5 and obtains a set S of 5 nearest neighbor vectors of R, where the positive examples (diagnosed as cardiovascular patients) form a set S1. The predicate Divide calculates the quotient |S1|/K which is the chance (possibility) that Qian has got cardiovascular disease.

We will run this program in next sub-section on a large set of patients' data and check the preciseness of results.

### 3.4 Qualified Efficiency of Vector Programming

When testing the 2000 sample data left in sub-section 3.2, we first used algorithm 3.2 to calculate a virtual vector for each data item with J = 5. Then we used its K-nearest neighboring vectors to estimate the label (yes or no) of this virtual vector with K = 10. The results showed that the average correctness rate of the 2000 samples' test was 70.80%.

In order to compare the success rate with other methods, we used the Logistic regression model [20] to do the same job. For preprocessing the data, we took the tail terms (i.e. the data part) of all triplets as Logistic training data and the labels Yes/No as 1/0 parameters. After preprocessing we used the open source Logistic regression routine [21] to test these data where the super parameters solver = sag, random_state = 3000, max_iter = 600 and other parameters were default. Finally, we got the precision rate 70.25%. This shows that our result produced by vector space representation and TransMETH algorithm is better.

**Experiment 3.3:** To further estimate the advantage of vector space representation and TransMETH algorithm, we select 10 positive and 10 negative samples randomly for checking the preciseness of the K-nearest neighbor method on a series of layered K-values. Theoretically, it is clear that the smaller the K value is, the closer should be the K-neighboring vectors to the central vector, and (as a conclusion) the more precise will be the estimation of central vector's label (yes or no) with the average value of k-nearest vectors' labels.

Execute the following Query:

**Query:** ? cardiovascular (Data (Patient), Cardi-V, 10, R)

**Rule:** cardiovascular (Data (X), Cardi-V, J, R) ←
　　　　Construct (Data (X), Cardi-V, J, R), Nearest (R, Cardi-V, K, S),
　　　　Positive (S, S1), Divide (Size (S1), K, Chance), Fail (K).　　　　(3-5)

**Data:** Patient: {Positive-i | 1 <= i <= 10}, {Negative-i | 1 <= i <= 10}.

**Data:** K: 5,10,20,30,40,50.

**Note 2:** For doing the experiment, 20 patients data were randomly fetched from the database Cardio, where 10 patients are positive (suffering from cardiovascular disease) and 10 were negative.

The results are shown in Table 3-1 and Figure 3-1.

Table 3-1　Result of 10 patients suffering from cardiovascular disease

| sample<br>K value | 1 | 2 | 3 | 4 | 5 | 6 | 7 | 8 | 9 | 10 | Average |
|---|---|---|---|---|---|---|---|---|---|---|---|
| 5 | 0.8 | 0.6 | 1 | 1 | 0.8 | 0.8 | 0.2 | 0.4 | 0.8 | 1 | 0.74 |
| 10 | 0.9 | 0.7 | 1 | 1 | 0.5 | 0.7 | 0.3 | 0.4 | 0.8 | 0.9 | 0.72 |

| 20 | 0.8 | 0.45 | 0.8 | 0.9 | 0.6 | 0.7 | 0.2 | 0.3 | 0.65 | 0.7 | 0.61 |
| 30 | 0.67 | 0.4 | 0.77 | 0.93 | 0.7 | 0.73 | 0.27 | 0.3 | 0.6 | 0.67 | 0.604 |
| 40 | 0.7 | 0.35 | 0.82 | 0.9 | 0.68 | 0.75 | 0.25 | 0.3 | 0.57 | 0.6 | 0.592 |
| 50 | 0.72 | 0.32 | 0.78 | 0.9 | 0.64 | 0.8 | 0.2 | 0.36 | 0.6 | 0.58 | 0.59 |

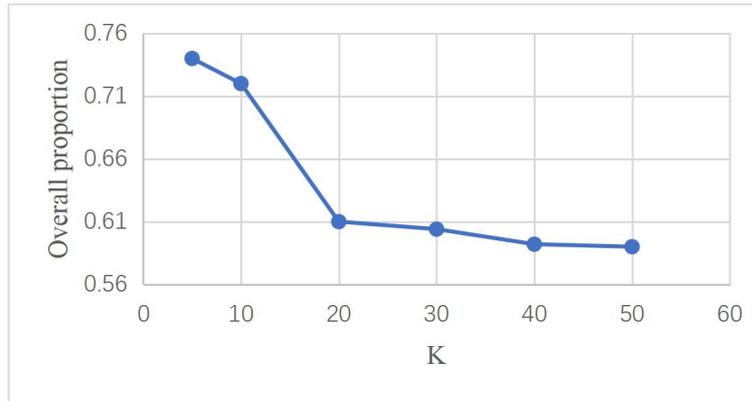

**Figure 3-1**　Calculated chance variation of 10 patients suffering from cardiovascular disease

**Note 3:** The general tendency of average chance variation is correct: The larger the K-value is, the less precise is the average estimation of positive examples, since the larger the K-value is, the more negative examples will be mixed into the K-circle.

**Note 4**: Sample 7 and sample 8's estimated chance values were relatively low. This is because for each of them most attribute values were quite good but one attribute's value (systolic blood pressure, 190)) was very bad.

Table 3-2 and Figure 3-2 show the similar tendency for negative samples.

**Table 3-2**　Results of k-nearest Neighbors of Negative Samples

| sample<br>K value | 1 | 2 | 3 | 4 | 5 | 6 | 7 | 8 | 9 | 10 | average |
|---|---|---|---|---|---|---|---|---|---|---|---|
| 5 | 0.6 | 0.8 | 0 | 0.4 | 0.6 | 1 | 0.4 | 0.6 | 1 | 0.8 | 0.62 |
| 10 | 0.5 | 0.7 | 0.1 | 0.5 | 0.7 | 0.9 | 0.6 | 0.4 | 0.8 | 0.7 | 0.59 |
| 20 | 0.5 | 0.4 | 0.2 | 0.4 | 0.6 | 0.8 | 0.6 | 0.35 | 0.75 | 0.6 | 0.52 |
| 30 | 0.5 | 0.4 | 0.13 | 0.4 | 0.63 | 0.8 | 0.63 | 0.27 | 0.77 | 0.67 | 0.52 |
| 40 | 0.47 | 0.38 | 0.12 | 0.35 | 0.6 | 0.78 | 0.55 | 0.28 | 0.8 | 0.65 | 0.498 |
| 50 | 0.46 | 0.36 | 0.14 | 0.32 | 0.66 | 0.78 | 0.5 | 0.3 | 0.8 | 0.64 | 0.496 |

**Note 5:** The attributes' values of sample 3 were not good. Therefore, there were many positive samples calculated around him (her). But in the database sample 3 was denoted as a negative example. This is why his (her) negativity was denied in the calculation.

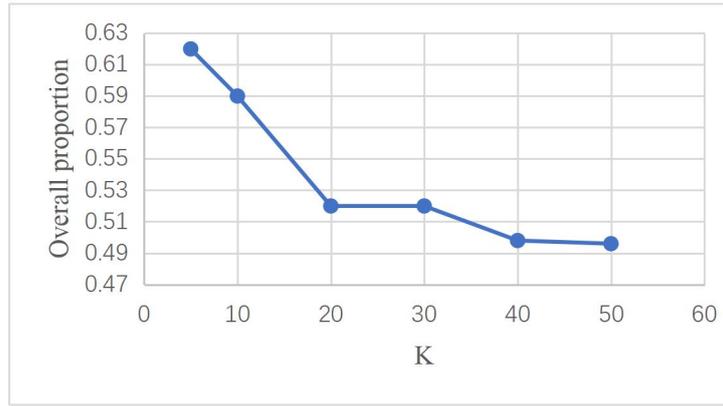

**Figure 3-2**  Calculated chance variation of 10 patients not suffering from cardiovascular disease

### 3.5  Shared Memory based Concurrent KG Embedding

In this sub-section we will introduce TransCMETH, which is a shared memory concurrent version of the TransMET algorithm, where C means concurrency. TransCMETH is a further development of TransMET, where some new ideas come from the reference [78-79]. The techniques of concurrent KG embedding proposed in the literature may have good effect because they are usually used on the large KGs such as DBpedia or Yago. Since such large KGs contain a large quantity of knowledge in different areas, the vectors constructed by these algorithms are usually sparse in the whole vector space. It follows that the plausibility that the parallel threats mine the same sample data at same time is low [80-81]. This is the reason why such algorithms usually have less conflicts (Different data embedded to approximate vectors, or contradicting data embedded to approximate vectors). The convergence of optimization algorithms is rarely affected. However, in our case the situation is different. Take the Cardi KG as example. Its knowledge consists of a huge set of data containing only domain knowledge of a very limited amounts of types. This characteristic would produce much worse effects on algorithm convergence.

In order to lower down these bad effects, we introduce two improvements in the new algorithm TransCMETH, where the letter C means concurrency, H means hash technique. On the one hand, we use a hash table to store and fetch data values before and after vectorization for improving the search efficiency. On the one hand, for each pair of positive/negative samples $((t_1, t_2, ..., t_m, l), (t_1', t_2', ..., t_m', l'))$ we only renew the triplet's vector (the $t_i$s) of positive samples, while we let the negative samples unaffected. On the other hand, while optimizing the vectors their components are not necessarily optimized by all threats at the same time. Each thread may select a part of the vector components for optimization. These two measures will help the new algorithm lower down the possibility that two threats try to optimize two vectors at the same time..

---

**Algorithm 3.3** TransCMETH

**input** Training set S={ $(t_1, t_2, ..., t_m, l)$ }, weights set W={$w_1, w_2, ..., w_m$}, where we call each $(t_1, t_2, ..., t_m, l)$ an m+1 tuple. The $w_i$, where all $w_i > 0$, $1 \leq i \leq m$, $\sum_i w_i = 1$, is the weight of $t_i$ respectively. Let L be the labels set, and $T_i = \{t \mid t$ is a triplet of i-th kind$\}$, i=1,2,…,m. $\gamma$ is the margin, p is the number of threads (processes), $p_j$ is j-th thread (processor), $1 \leq j \leq p$. $n$ is the embedding dimension of $t_i$ and $l$.

1: **initialize** $l \leftarrow$ uniform$(-\frac{6}{\sqrt{n}}, \frac{6}{n})$ for $l \in L$ // Persist in the shared memory [78]

2: $\quad\quad\quad t_i \leftarrow$ uniform$(-\frac{6}{\sqrt{n}}, \frac{6}{\sqrt{n}})$ for $t_i \in T_i$ // persist in the shared memory

3: **loop**

4: $\quad\quad t_i \leftarrow t_i/\|t_i\|$ for each triplet $t_i \in T_i$

5: $\quad\quad l \leftarrow l/\|l\|$ for each label $l \in L$

6: $\quad\quad$ For each of the p threats $p_j$ do $\quad\omega$ In Parallel [78]

7: $\quad\quad\quad SP_{batch_j} \leftarrow$ sample $(S(p), \frac{b1}{p})$ // sample a minibatch of size $\frac{b1}{p}$ from positive examples of S.

8: $\quad\quad\quad SN_{batch_j} \leftarrow$ sample $(S(n), \frac{b2}{p})$ // sample a minibatch of size $\frac{b2}{p}$ from negative examples of S, Where $S(p) \cup S(n) = S$, b1 = c|S(p)|, b2 = c|S(n)|, 0 < c <1.

9: $\quad\quad\quad Q_{batch_j} \leftarrow \emptyset$ // initialize the set of pairs of m-triples

10: $\quad\quad\quad$ for $(t_1, t_2, ..., t_m, l) \in SP_{batch_j} \cup SN_{batch_j}$ do

11: $\quad\quad\quad\quad (t_1', t_2', ..., t_m', l') \leftarrow$ corrupt$((t_1, t_2, ..., t_m, l), k)$ // Replace k $t_i$s with k $t_i$'s and $l$ with $l'$ for each m+1 tuple in $SP_{batch_j} \cup SN_{batch_j}$, where k is randomly selected each time. $l'$ is the opposite of $l$.

12: $\quad\quad\quad\quad Q_{batch_j} \leftarrow Q_{batch_j} \cup \{((t_1, t_2, ..., t_m, l), (t_1', t_2', ..., t_m', l'))\}$

13: $\quad\quad\quad$ End for

14: $\quad\quad\quad$ Update embeddings w.r.t.

15: $\quad\quad\quad V_j \leftarrow$ sample $(n, v)$ // sample a set of size v from $(0, 1, …, n-1)$

$$\nabla_{V_j}[\gamma + d(w_1 t_1 + w_2 t_2 + ... + w_m t_m - l) - d(w_1 t_1' + w_2 t_2' + … + w_m t_m' - l')]_+$$

// only the dimensions of $t_1, t_2, ..., t_m, l$ that appear in the $V_j$.

16: $\quad\quad$ Store all regulated addresses of combined m+1-vectors in the HH $((w_1 t_1, w_2 t_2, ..., w_m t_m, l))$ address of the shared memory-based hash table.

17: $\quad\quad$ If one of threads not finished do

18: $\quad\quad\quad$ wait(0.01 second) // wait for all sub-threads finish

19: $\quad\quad$ End if

20: **End loop**

**Experiment 3.4.** We performed experiment 3.1 once again by using algorithm TransCMETH instead of TransMET. The results are shown in Table 3-3 and Figure 3-3, where the number of threads is 1, 2, 4, 8 and 16. The experiment was done three times for each number of threads. The values in the table 3-3 are the average result of three times of computation of each threat's number. It can be seen that the results' preciseness decreases slowly from 0.7080 to 0.6875, while the efficiency improvement increases from 1.00 to 6.25. We use the following equation (3-6) to estimate the gain this algorithm has brought:

Algorithm Gain = Loss of Preciseness × Gain of Efficiency
$\quad\quad\quad$ = (0.6875/0.7080) × (6.25/1.00) = 6.069032485875707 $\quad\quad\quad\quad$ (3-6)

The result of experiment 3.4 is shown in Table 3-3 and Figure 3-3.

**Table 3-3** Results of Experiment 3.4

| Number of Threads | Preciseness | Used Time in Seconds | Speedup Ratio |
|---|---|---|---|
| 1 | 0.7080 | 3271 | 1.00 |
| 2 | 0.7008 | 2054 | 1.59 |
| 4 | 0.6992 | 1136 | 2.88 |
| 8 | 0.6885 | 634 | 5.16 |
| 16 | 0.6875 | 523 | 6.25 |

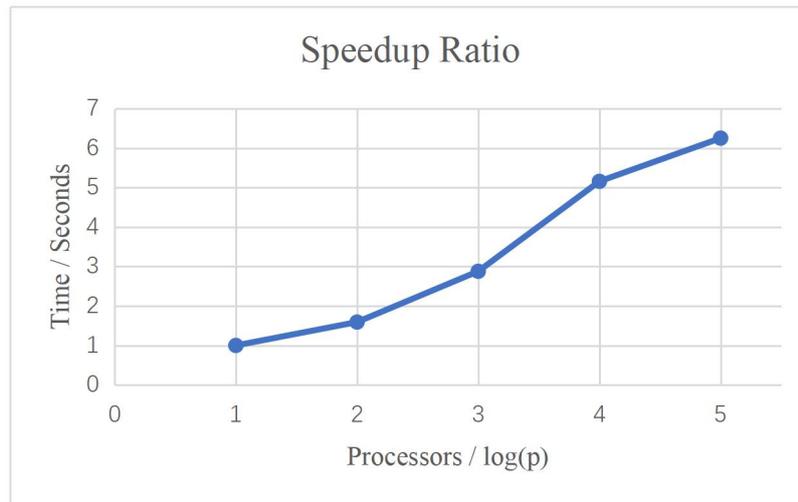

**Fig.** 3-3 Speedup Radio of Experiment 3.4

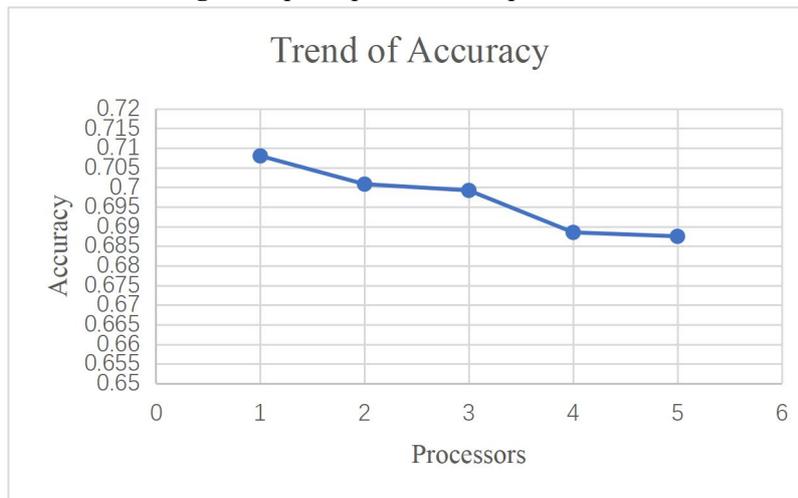

Fig. 3-4 Decrease in Accuracy of Experiment 3.4

Discussion: By considering Table 3-3 and Figure 3-3 we will see that with the exponential increase of threads number the speed-up ratio first increases exponentially and then only logarithmically. This tendency of variation is understandable. It tells that the speed-up rate will increase effectively if the number of threads is kept on a rational level. Whenever the number of threads exceeds a certain threshold value, the corresponding speed-up ratio is no more remarkable due to the limitation of computation resources. As for the decrease of result's preciseness with increase of threads number, the reason is the rise up of possibility that the threads suffer from collision with each other with the boosting of threads generation.

# § 4 Unsharp and Uncertain KGN-Prolog Programming

In this paper, by 'unsharp' we mean 'fuzzy', and by 'uncertain' we mean 'probabilistic'.

## 4.1 A Short Review of Unsharp Logic Programming

Historically, Zadeh introduced fuzzy set and fuzzy systems in 1965 [22], where the fuzziness of a system is characterized by a digit between 0 and 1. Later he introduced an advanced fuzzy set concept in 1975 [23], where a system's fuzziness can be characterized by words (concepts). Therefore, this kind of fuzziness is called by researchers as computing with words [24]. But they have also got another name each. The digital measure based fuzziness has been called type-1 fuzziness, whereas the fuzzy language based one called type-2 fuzziness. Correspondingly, people talked about type-1 resp. type-2 fuzzy sets and fuzzy systems.

The advancement of fuzzy set and fuzzy systems' study led also to an intensive research on introducing fuzziness in logic generally and in logic programming language specifically. In particular, the fuzziness concept was introduced in Prolog language and quite a lot of fuzzy Prologs have been invented, of which some ones are based on type-1 fuzziness, and some others based on type-2 fuzziness. In this paper we only care the application of these researches in designing new fuzzy Prolog versions and would like to call the former as type-1 fuzzy Prolog and the latter as type-2 fuzzy Prolog, although we did not find such renaming in literature.

Most of the early type-1 fuzzy Prologs define fuzziness as discrete values in the closed interval [0, 1] and use Min/Max operations to implement and/or fuzzy calculation. For example in Prolog-ElF [25] the truth values are divided in 11 levels from 1.0 to 0.0, However, only truth values between 0.5 and 1.0 could be rationally processed, because [26] has proved that the resolution principle it used is only guaranteed to be meaningful if all truth values to be processed are no less than 0.5 (with zero as exception). [27] has improved this result to allow the truth value to be in the closed interval [0, 1]. Another example of such systems is FProlog [28], where each predicate in the rule body has a fuzzy value. Later the Fuzzy Prolog of [29] made a step forwards by proposing the fuzziness defined as the union of some sub-intervals within the interval [0, 1], where some aggregation operators are performed over the combination of sub-intervals when different evidences about the same subject are collected.

On the other hand, Bousi~Prolog, BPL for short, first published in 2009 [30-31], is a type-2 Prolog based on dictionaries such as Wordnet::Similarity [32]. The inference makes use of fuzzy unification based on proximity of words (concepts), which is a binary fuzzy relation defining a fuzzy subset of words, where the proximity of a pair of descriptive words is defined as a real number within the closed interval [0, 1]. These words are not limited to predicates, but can also be function names and parameters. A further development of Bousi~Prolog is the fuzzy logic programming language FASILL[33-34] that integrates Bousi~Prolog with FLOPER (Fuzzy LOgic Programming Environment for Research), where a notable property of it is, besides proximity, that a partial ordering in form of a complete lattice of words is incorporated.

Proximity, or similarity, based fuzzy computing has been used for enriching the Prolog language semantics. [35] describes Rfuzzy, a fuzzy Prolog language of which the fuzzy reasoning is lattice theory based sub-interval truth value calculation (union, intersection, etc.) and propagation. According to this logic, a truth value is a finite union of sub-intervals on [0, 1]. Truth values can be calculated by conjunction and/or disjunction. These sub-intervals build a Boolean algebra which forms a complete lattice. In this logic crisp and fuzzy reasoning can work together smoothly. The authors also provided declarative and procedural semantics of this Prolog.

The Rfuzzy version proposed by [36] can be considered as a light variant of [35], where truth values are simply real numbers rather than sub-intervals. It has a Prolog like syntax. Furthermore, it provides clauses with default fuzzy values, partial default values and predicates with types.

While in the approaches proposed by [35]and [36] each concept is characterized by a single fuzzy index, [37] introduced a multiple indices approach for differentiating concepts based on a value lattice. This technique was used to calculate preferences among various alternatives. For example, assume the doctors are going to take an operation on some patient. There are different indices for evaluating different operation programs, including safety, prognosis, painfulness and costs. Then different combinations of these indices' values form a lattice for each patient.

### 4.2  A Short Review of Uncertain Logic Programming

The earliest probabilistic logic programming language we found was proposed in [38]. In this language each predicate is attached with a probability represented as a sub-interval of [0,1]. Thus [0.3, 0.7] : P (a, b) means the probability of P (a, b) is between 0.3 and 0.7. On the other hand, [0, 0]: P (c, d) means the negation of P (c, d). Beside the probability, its main difference from standard Prolog is mainly in two aspects. The first one is the lack of function symbols. This makes it closer to Datalog than to Prolog. The second one is the admittance of disjunction of predicates in rule's body. Another proposal made by [39] suggested to introduce conditional constraints in probabilistic logic programming. Different from the above ideas, the probabilistic Prolog language ProbLog proposed in [40] assigns a probability to each clause. This probability does not mean the degree of truth of each rule, but the chance of each rule to be selected for inference each time whenever more than one clause can be called for at run time. Its second version, ProbLog2, integrates the idea of probabilistic programming with statistical relational learning techniques [41]. Another probabilistic logic programming language is PRISM, which is similar to probabilistic Prolog but has a capability of learning from examples [42].

The combination of probability and fuzziness uncertainty has also been challenged, for example in the relational language Fril in form of probabilistic fuzziness [43]. But we will not discuss this direction of research here since it may cause semantic confusions.

### 4.3  Introduce type-3 fuzzy-probabilistic logic programming

In this sub-section we introduce the type-3 fuzzy probabilistic KGN-Prolog language. Or more exactly, the type-3 fuzzy probabilistic Prolog, which is based on partial orders. Different from other partially ordered objects of fuzzy Prologs, it has the following significant characteristics, where the antonym of fuzzy is crisp:

**Definition 4.1.**

1) A type-3 partial order is a mixed concept-digit partial order where each element is a n-tuple of [digits or concept]. The number n is fixed for the whole partial order

2) Element x is less or equal than element y if each component of x is less or equal to the corresponding component of y.

3) Any element X without explicit n-tuple is assumed to have an implicit or unknown n-tuple satisfying the above rule 2.

4) It may be interpreted as a pure fuzzy or a pure probabilistic lattice upon programmer's request. Fuzziness and probability should not be mixed in the same lattice.

5) (1) Any fuzzy (probabilistic) value (in a n-tuple) belongs to the interval [0, 1]. (2) The top (bottom) value for fuzziness and probability is 1 (0). (3) Both kinds of values are increasing monotonically on each path from bottom to top.

6) In each rule of a type-3 fuzzy (probabilistic) KGN-Prolog program, a fuzziness (probability) tuple can be attached to any predicate of the body part. Predicates without fuzziness (probability) attachment have the default crisp (deterministic) value 0 (1) for all elements in the n-tuple.

7) The runtime fuzziness (runtime probability) of the rule head is calculated as a tuple of which each element is the maximal (product) value of its own dynamic fuzzy (probabilistic) value with those of all its body predicates.

8) There are different semantics of partial orders consisting of fuzzy (probabilistic) concepts. These semantics can display their static partial orders, or their monotonic clarification process, or their stepwise evolution in the procedure of some event process. These differences will lead to different logic programming styles.

**Example 4.1.** Considering reference [73] of larynx cancer therapy, where the structure of different therapies forms a partial order where the fuzziness tuple is a binary one. The digits pair shows each therapy's prognosis with the tuple (cure rate, voice quality). In Figure 4-1 we establish a type-3 fuzzy partial order of this example where the three concrete therapies 'take out', 'radioactive therapy' and 'hemi-gectonomy' are replaced with three abstract ones A, B and C. Their assumed estimation values may come from some statistics. The arrows show the 'less or equal than' direction of the partial order.

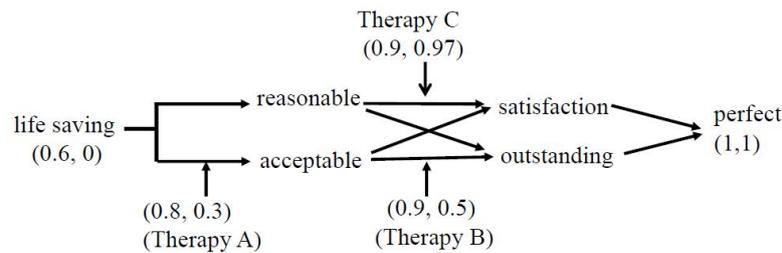

**Figure** 4-1. A type-3 partial order of larynx-therapy effects

A complete different semantics of partial order is the temporal evolution of fuzzy (probabilistic) evaluation of events or situations. When hospital doctors check a patient's disease, they may get some fuzzy impression about the disease and/or probabilistic expectation of its development (after drug taking or operation) at the beginning. Such impression and expectation will change dynamically with the time due to more and more medical tests and treatment effects. Such kind of reasoning process may be represented by a temporal partial order.

**Example 4.2.** Investigate a more complicated example: Given a record [44] of consultation procedure about a patient who has got two plagues: on the one hand he displays a phenomenon of gluttony and on the other hand he has got problems of abnormal kidney function. After a detailed examination the doctors decided that this patient has a double problem of high glucose and a certain degree of kidney disease, including ERY and Pro. The doctors concluded that it is highly possible that the patient has got both CKD and diabetes problems. Finally, they decided on an exact conclusion: this is a typical case of diabetic nephropathy. By considering the fuzzy concepts occurring in the discussion process, we discover a hierarchy of fuzzy concepts, which displays the stepwise evolution of a medical consultation, where doctors discuss and explore the patient's condition for deciding the disease. The patient has got tumor at forehead and in both lungs. Figure 4-2 shows the fuzzy reasoning process with both the reasoning concepts and their fuzziness degrees.

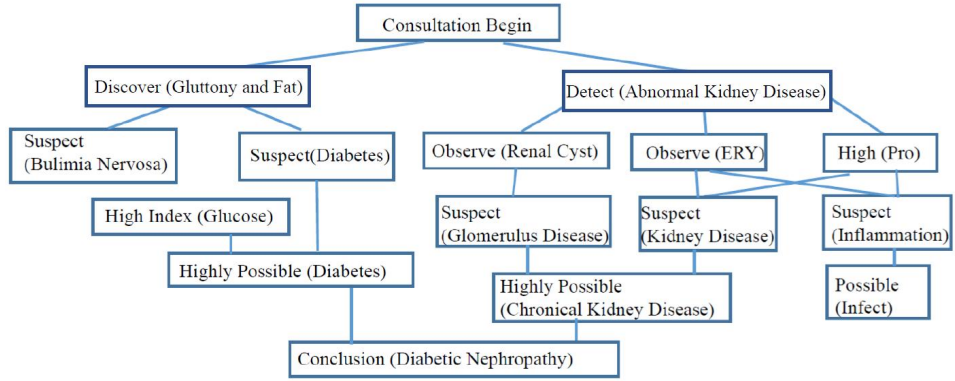

**Figure** 4-2. A Partial Order of Fuzzy Reasoning

**4.4   Fuzziness, Partial Order and Comparative Unification**

The theoretical basis of Prolog is Horn logic which consists of a finite set of Horn clauses. Each Horn clause is a finite disjunction of predicates, where at most one predicate is positive. All others must be negative. Thus, the Horn clause $P \vee \neg P_1 \vee \neg P_2 \vee \neg P_3 \ldots \vee \neg P_n$ represents the rule $P \leftarrow P_1, P_2, P_3, \ldots, P_n$. Using the unification and resolution technique found by Robinson one can get a high efficiency procedure of solution finding. In this article, we also introduce a generalized form of unification, the comparative unification, which allows the unifier making use of semantic information when processing some predicates for relative comparison, such as Eq, Larger, Smaller, etc. We first consider an example:

**Example 4.3:** Consider the following rule, where the predicate Larger (age (X), 70) is a semantic predicate.

**Rule**:  Send-to (X, ICU) ← Get-disease (X, ABC), Larger (age (X), 70).

**Data**:  Eq (age (Wang), 75).

Consider the query ? Send-to (Liang, ICU). The question is: should the doctor send Liang to ICU? It is impossible to unify Larger (age (X), 70) with Eq (age (Liang), 75) in the usual sense. The standard definition of resolution does not apply in this case. But the answer should be 'yes' since for any X, Larger (age (X), 70) is the consequence of Eq (age (X), 75) for any X. In order to solve this problem, we introduce a new kind of them can be unified in a generalized way to get Larger (age (Wang), 70). However, this result cannot be derived from the resolution principle, because the standard unification principle does not unify two predicates with different names. For that purpose we should have a new kind of unification: the comparative unification.

Rewriting a triplet(X, Y, Z) in Y (X, Z) form is to transform it in predicate form such that during Prolog program running the comparative unification process can be performed.

**Definition 4.2** Predicate P is said to be comparatively unifiable to Q if and only if

1) There is a maximal general unifier (MGU) f transforming some common variables of both sides such that f(P) can be used to validate the truth of f(Q).

2) If Q is a predicate in the body part of a rule R and P is a datum or the head part of another rule W, then f (R) [f (P)/f(Q)] is a valid rule where f(Q) is replaced by f(P).

3) We call such unification a comparative unification, the predicate being unified a comparative predicate.

The principles differentiating comparative unification from others include:

1) It is not based on proximity, but on partial ordering of predicates.

2) The partial ordering can have different semantics (e.g. in the sense of numerical comparison, of logical implication, or of individual preference, etc.).

3) Partial ordering is the only deciding factor of unification. Other things, such as the parameter structure of predicates, do not play a role (e.g. classical unification can never unify Larger (age (X), 70) with Eq (age (Liang), 75)).

4) It is one-sided (predicate P unified to Q), rather than bi-sided (P and Q are unified).

5) It is a transitive relation, but proximity isn't.

6) It isn't a symmetric relation, but proximity is.

Introducing comparative unification has an additional advantage. It makes the use of KG's knowledge easy. A stipulation of KGN-Prolog is the permission of representing a KG triplet (A, B, C) in form of a predicate B (A, C) during comparative unification. For example, the triplet (heartbeat (Zhang), eq, 83 beats/minute) functionally equals to the predicate eq (heartbeat (Zhang), 83 beats/minute). This stipulation makes the direct use of KG triplets in logic reasoning easy and comfortable.

**Definition 4.3**: The second term B of a triplet (A, B, C) is called a comparative relation if B (A, C) can be used as a comparative predicate. The triplet itself is called a comparative predicate.

The KGN-Prolog system owns a library of semantics predicates with corresponding semantic unification algorithms.

## § 5  KGN-Prolog for Medical Image Analysis

**5.1  Image Processing and Beyond: The Multimodal Applications of Prolog**

Medical image processing has been playing a more and more significant role in applying AI and image processing techniques to the medicine and healthcare domain, including image analysis, classification, segmentation, mining, etc. Many large sets of medical images of different sorts need to be processed and benefited, where the dimensions of images have been increasing steadily. 'The radiology image matrices may vary from 64×64 for some nuclear medicine exams, to over 4000×5000 for some mammogram images'[45]. Modern requirement of medical image processing appeals to more advanced technology for processing them efficiently. On the other hand, profound studies on processing these image data sets need more high quality medical data sets, in particular the public ones. Only seven years ago, in 2016, people were complaining that "everyone participating in medical image evaluation with machine learning is data starved" [45].

This situation appeared to be improved recently. It was reported in [46] about some large public datasets including the ChestXray14 (CXR14) dataset with over 112,000 chest radiographs [47] and the Musculoskeletal Radiology (MURA) dataset containing over 40,000 upper limb radiographs3 [48].

Not limited in image processing only, Prolog languages have been used in other multimodal areas. One of these aspects is the NLP technique. Just an example: the Quintus Prolog has been applied in the ShopTalk system in manufacturing area [49], where it not only handles images and graphics directly, but also processes natural language texts describing complex picture scenes. [50] even extended the traditional definite clause grammar of Prolog to include multimodal definite clauses for accepting real-time natural language instructions. [51] built a multimodal interface supported by Prolog, which is able to understand voice speech and handwriting inputs. The most interesting works in this direction may be the Prolog programming in robotics. As early as in 1986, multi-modal techniques have been introduced in Prolog2 for processing not only images, but also for camera control, illumination sources, optics, video and even simple robotic devices [52]. Recent advances in this direction include [53], which provides real-time aids to disabled persons with natural interaction in an open environment where almost every event can occur without a predefined list of services.

**5.2   KGN-Prolog Application in Image Analysis**

As an example of KGN-Prolog's application power, we introduce in this section the image processing facility to its programming, including image classification and retrieval. Its core techniques include image recognition and classification. As it is well-known, the most successful machine learning technique for image analysis is the convolutional neural network [54]. As a typical example we mention the Multigrain structure developed in [55] which can recognize images based on object class, particular object, or distorted copies of the same object. In the medical domain, this technique can be used to recognize different sorts of diseases, different kinds of images (X-ray, MR, B-US, etc.), and different personalities (Lang term investigation of the same patient). Moreover, this technique can be used to compare a new patient's image with similar past images in the database to get a first impression about what has happened with this new patient. In order to assign image retrieval function to the usual image classification framework, the Multigrain structure introduced an additional final pooling layer and the whitening technique.

For displaying the image processing functions of KGN-Prolog, we have made the following two experiments done at our lab.

**Experiment 5.1**. We first tried the classical methods like [55] without Prolog application. But different from Multigrain, we only used a general convolutional neural network to embed images in vector spaces. For image retrieval we used the vector space operation technique introduced in section 3 of this paper. More exactly, our idea is first to use a convolutional neural network to transform each image to a vector and then to develop logic programming on the space of these vectors. This general structure is roughly shown in Figure 5-1, where the two images were taken from [48].

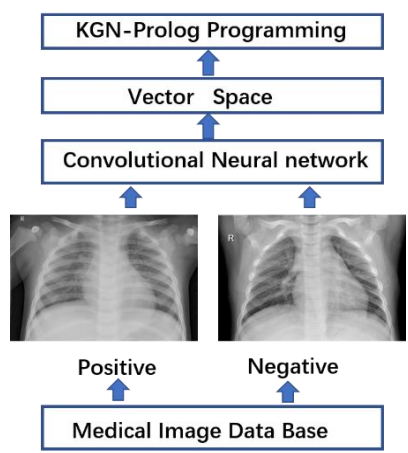

**Figure** 5-1 KGN-Prolog Programming on Image Data Bases

In order to apply the medical image processing capability of convolutional neural networks, we have applied the pre-training model C2L [56] for training and testing our data— 14 subsets of the ChestXray data set. This model is based on the classical ResNet-18 architecture [57].  The super-parameters used in training are epoch=120，learning rate =0.03，batch size = 128, respectively. As a result, we got the average preciseness 83.5%.

**Experiment 5.2**. On the other hand, we have also used our k-nearest neighbors technique developed in section 3 to test the same database. For that purpose, we have enriched KGN-Prolog with a built-in function cnn (Med, X, V) working on the vector space, where Med is the name of algorithm used, e.g. c21. X is the sample and V is the result vector of calculating X.

In this experiment, we select two subsets Pneumonia and Edema from [48] as (positive) training data. The k-nearest neighbor results of Pneumonia are shown in Table 5-1, where the digits in the table tell 'How much is the percent of positive ones among all neighboring k vectors'. It shows that this percent is monotone decreasing with increasing k-values. The corresponding diagram is shown in Figure 5-1.

**Table** 5-1 Results of Generating k-nearest Neighbors of Positive Samples of Pneumonia

| Sample / K value | 1 | 2 | 3 | 4 | 5 | 6 | 7 | 8 | 9 | 10 | Average |
|---|---|---|---|---|---|---|---|---|---|---|---|
| 5 | 0.8 | 1 | 0.6 | 1 | 0.4 | 1 | 1 | 0.8 | 1 | 1 | 0.86 |
| 10 | 0.8 | 1 | 0.6 | 0.9 | 0.5 | 0.9 | 0.9 | 0.9 | 0.9 | 1 | 0.84 |
| 20 | 0.85 | 1 | 0.6 | 0.85 | 0.45 | 0.85 | 0.95 | 0.85 | 0.9 | 1 | 0.83 |
| 30 | 0.83 | 0.93 | 0.63 | 0.83 | 0.47 | 0.87 | 0.97 | 0.87 | 0.93 | 0.93 | 0.826 |
| 40 | 0.82 | 0.9 | 0.68 | 0.8 | 0.45 | 0.85 | 0.97 | 0.85 | 0.9 | 0.95 | 0.817 |
| 50 | 0.84 | 0.88 | 0.68 | 0.8 | 0.44 | 0.86 | 0.96 | 0.8 | 0.88 | 0.96 | 0.81 |

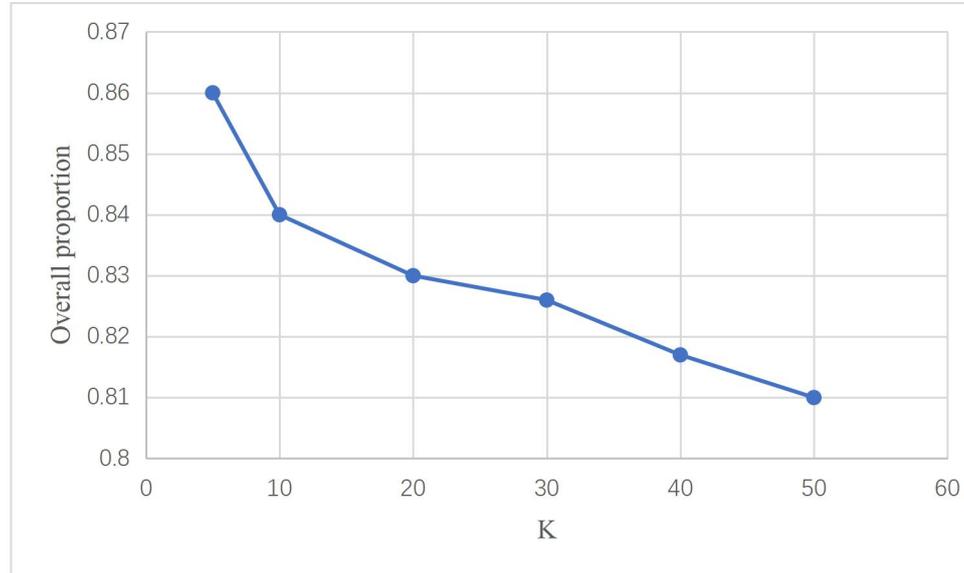

**Fig.** 5-1 Decrease of Pneumonia's Positive Neighbors' Percent with Increasing k Values

The k-nearest neighbor results of Edema are shown in table 5-2, where the digits in the table tell 'How much is the percent of positive ones among all neighboring k vectors'. It shows that this percentage monotonically decreases with increasing k. The corresponding diagram is shown in **Figure 5-2**.

**Table 5-2** Results of Generating k-nearest Neighbors of Positive Samples of Edema

| Sample / k-value | 1 | 2 | 3 | 4 | 5 | 6 | 7 | 8 | 9 | 10 | Average |
|---|---|---|---|---|---|---|---|---|---|---|---|

| 5 | 1 | 1 | 1 | 1 | 1 | 1 | 1 | 1 | 1 | 1 | 1 |
|---|---|---|---|---|---|---|---|---|---|---|---|
| 10 | 1 | 1 | 1 | 0.9 | 1 | 1 | 1 | 1 | 1 | 1 | 0.99 |
| 20 | 1 | 1 | 1 | 0.95 | 1 | 0.95 | 1 | 1 | 1 | 1 | 0.99 |
| 30 | 1 | 1 | 1 | 0.97 | 1 | 0.97 | 0.97 | 1 | 1 | 1 | 0.991 |
| 40 | 1 | 1 | 1 | 0.97 | 1 | 0.97 | 0.97 | 0.97 | 1 | 1 | 0.988 |
| 50 | 1 | 0.98 | 1 | 0.98 | 1 | 0.96 | 0.98 | 0.98 | 1 | 0.98 | 0.986 |

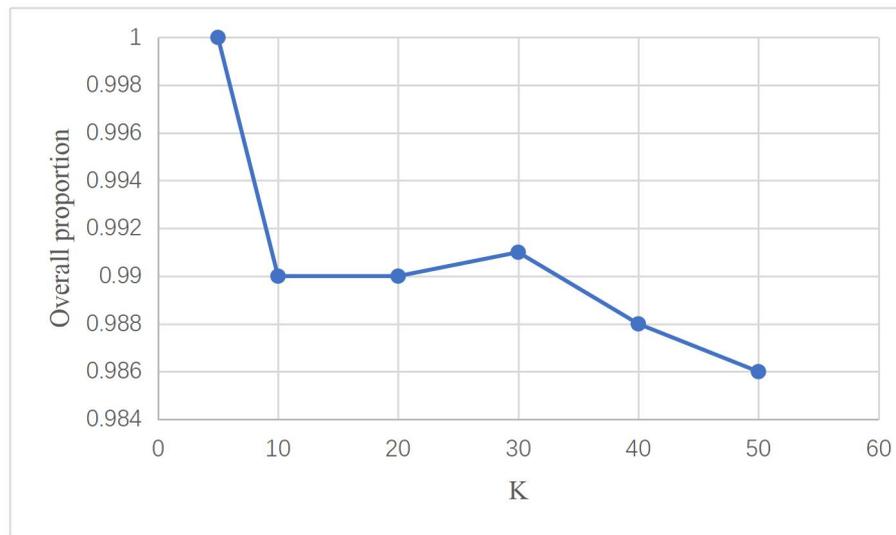

**Fig**. 5-2 Decrease of Edema's Positive Neighbors' Percent with Increasing k Values

## $ 6    KGN-Prolog for Consultation and Argumentation

**6.1    Machine aided Medical Consultation and Argumentation**

In some sense, consultation is the most significant step in medical decision making. Correspondingly, research on machine-aided (or AI supported) consultation became a significant branch of AI in medicine. In this sub-section we make a short summary of current directions in this area.

**1.    Argumentation as a protocol-based decision making process**. These works usually use a big file of medical documents for summarizing useful rules, guidelines or ontologies for decision making [58-60]. A more advanced work [61] has developed a platform to support the analysis of experts' argumentation records. Given an argumentation record, they rewrite all participants' arguments into standard forms and sort them to get a proposed result.

**2.    Argumentation as explanation tool for complex medical decisions.** The complex algorithms of the EIRA (Explaining, Inferencing and Reasoning about Anomalies) system applied in ICU domain proposed by [62] were first replaced by argumentation inference process partly in [63] and further completely in [64] with a new tool arguEIRA, which is an argumentation-based justification system.

**3.    Argumentation as an interactive planning support process.** [65] developed an argumentation based system REACT (Risks, Events, Actions, Consequences, over Time) for examining and evaluating patients treatment plan worked out by physicians in an interactive way where REACT and the physicians are the two sides of argumentation.

**4.    Argumentation as a medical students training tool.** Given a clinical problem without enough information for solving it. Medical students perform problem-based learning (PBL) by proposing hypotheses and trying to find solutions in group discussion [66].

    **5. Argumentation as an experts' discussion support process.** [67] used a whiteboard for decision making based on collaborative discussion with a structured template for constructing, annotating and sharing arguments.

**6.2   Argumentation as Preference Competition**

Among the various approaches of argumentation, argumentation as preference competition is the often used one. In this approach, the target of discussion is not to differentiate between true and false, but to differentiate between better and worse. Such kind of discussion appears also very often in clinical consultation. The doctors discuss about which therapy is best suitable for the current patient. On the other hand, patients also have their own wishes about which therapy or healthcare measure they want to accept.

Regarding some theoretical works in this direction, [68] introduced a game theory-based schema of interactive decision making by considering three prominent game structures. It is a theoretical model but can be of help for those building practical preference-based models. [70] extended the assumption-based argumentation frame work (ABA) [69] to a framework with preferences. The authors defined a set of decision criteria such that 'good decision' can be decided during argumentation by a given computation script.

Regarding technical studies of preference enhanced-argumentation we mention [71] which introduced a decision support system based on a theory of threefold stakeholders: information collected from medical tests, patient's preferences and doctors' explanations. The authors have built a platform Consult for supporting preference-based argumentation, where the preferences can be integrated into the system or proposed during system running for sharing by the stakeholders. On the other hand, [72] used a practical approach of resolving the mis-understanding problem between doctors and patients. It introduced the concept of Evidence-based Argumentation Graph which consists of a clinical argumentation scheme and a patient preference argumentation scheme. Each decision should meet a balance between doctor's and patient's preference.

Preference competition also plays a central role in KGN-Prolog's argumentation processing.

**6.3   Preference Competition Represented as Partially ordered Sets**

Let's first give a formal representation of argumentation used in this paper.

**Definition 6.1**. 1. Each argument is a bi-sectioned rule where each side (head part and body part) is a finite conjunction of predicates. It means: if all predicates of the body (rights) part are true then all predicates of its head (links) part are also true.

**2.** Each of its head (body) part may be empty, but not both.

**3.** If the head part is empty, it means the body part is false.

**4.** If the right side is empty, it means the head part is data (unconditionally true).

**Definition 6.2**. Each session of argumentation (simply: each argumentation) consists of a timed sequence of arguments, where each argument can be marked with the name of the debater optionally.

**Example 6.1**   Following is an argument with both head and body parts not empty:

    General anesthesia (X), Send-to-after (X, ICU) ← Major-surgery (X), Aging (X).    (6-1)

Definition 6.1 and 6.2 could be further generalized. For example, each of an argument's can be a disjunction of predicates instead of just conjunction. But in this paper we are limited to definition 6.1 and 6.2.

There may be two approaches of processing an argumentation. Either we consider it as a generalization of Prolog rules to non-Horn logic or we just consider it as a set of data to be processed by a KGN-Prolog program. In this paper we follow the second approach. The new thing we introduced here in KG-Prolog is partial ordering-based preference processing.

**Example 6.2** Given two alternative treatment regimens for a patient's larynx cancer. Each has its own advantage and disadvantage.

Cure rate (very good), Voice quality (low), Tolerate (very good if age < 75) ← Regimen1.    (6-2)
Cure rate (acceptable), Voice quality (good), Tolerate (bad for aging patients) ← Regimen2.

In order to decide between Regimen1 and Regimen2, we have to compare their left sides, i.e. their head parts. Which one is preferred? Regimen1 is better than Regimen2 in cure rate and tolerate, while Regimen2 is better in voice quality. The two Regimens are not comparable. To overwhelm the difficulty, we introduce the well-known criteria of comparing two sets according to their partial order.

**Definition 6.3**: Given two sets $s1 = \{u1, …, um\}$, $s2 = \{v1, …, vn\}$.    $m > 0, n > 0$.

1). If for each $ui \in s1$, there is $vj \in s2$, such that $ui <= vj$, then it is called an angelic partial order (i.e. Hoare partial order) set relation $s1 <A= s2$.

2). If for each $vj \in s2$, there is $ui \in s1$, such that $ui <= vj$, then it is called a demonic partial order (i.e. Plotkin partial order) set relation $s1 <D= s2$.

3). If both relations $s1 <A= s2$ and $s1 <D= s2$ are true, then it is called a complete partial order $s1 <AD = s2$.

In the above statements $<=$ ($<A=$, $<D=$, $<AD=$) means 'smaller or equal' (angelic $<=$, demonic $<=$, complete $<=$). Let's analyze the advantage and disadvantage of the three kinds of set partial orders. If we always prefer s2 whenever $s1 <A= s2$ in angelic partial order, then in case of $s1 = \{4, 6\}$, $s2 = \{2, 8\}$ we would have to accept the very low element '2' of s2. On the other hand, if we investigate the same example with demonic partial order, then we would have $s2 <D= s1$ instead. This means we would have to lose the very high element '8' of s2.

**Proposition** 6.1:

1). All three partial orders in above definition are reflexive and transitive.

2). All three partial orders in above definition are not symmetric.

Now we come back to example 6.2. Since each element (predicate) has a technical meaning, we would be only capable of comparing those predicate pairs with same technical meaning. In order to make the arguments in (6-2) comparable, we ignore the technical concepts 'cure rate', 'voice quality' and 'tolerability'. Then the set of prognoses for Regimen1 is {low, very good, very good if age < 75} (underlined in Figure 6-1). That of Regimen2 is {bad for aging patients, acceptable, good}(not underlined). For the moment we call them set1 and set2, respectively.

The partial order for these set elements is {low, bad for aging patients} <= acceptable <= good, and {good, very good if age < 75} <= very good. See Figure 6-1.

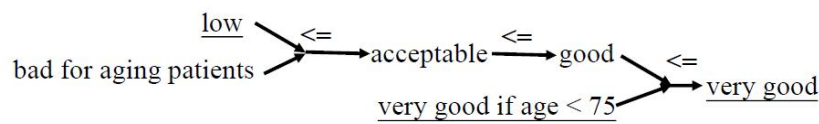

**Figure** 6-1 The <= partial order

We find that the relation 'Regimen1 <A= Regimen2' does not hold since the 'very good' element of Regimen1 does not equal or smaller than any element of Regimen2. The relation 'Regimen1 <D= Regimen2' does not hold neither, since no element of Regimen1 is equal or smaller than Regimen2's element 'bad for aging patients'. On the other hand, although the relation 'Regimen 2 <D= regimen 1' does not hold, but 'Regimen2 <A= Regimen1' holds.

**6.4    Partial Ordering based Arguments Analysis**

Of course, the pure partial order comparison introduced in the last sub-section is not a perfect solution. For example it does not consider the different weights of set elements. Generally speaking, in the context of larynx cancer therapy, cure rate is more important than voice quality.

In the following we will present a very simple model of preference-based arguments competition. In this model, all predicates don't have variables. All predicates are equally significant (There are no weights of predicates). Only the angelic partial order is considered. Each argument has only one predicate on its body side. These predicates are the competitors of the argumentation. The predicates on the head side of each argument form a set of characteristics of the corresponding competitor. The result of running this model is a partial order of all competitors organized by the angelic partiality relation.

**Example 6.3** As a target of discussion, we investigate the larynx therapy discussion of [73].

```
S1    (A1) My opinion is to take out the patient's larynx. This is has the best cure rate of 99%.
S2    (A2) I agree, taking out the patient's larynx would provide the best cure potential.
S3    (A3) I also agree, taking out the patient's larynx would provide the best cure potential.
RT1   (A4) But if you take out the patient's larynx, the patient will have no voice.
RT1   (A5) However, if you use radiotherapy, there is a 97% cure rate from the radiotherapy and about 97% voice
      quality, which is very good. The 3% who fail radiotherapy can have their larynx removed and most of these will
      be cured too.
S2    (A6) My opinion is also that the patient should have a hemi-laryngectomy. This will give a cure rate is as good as
      radiation therapy.
S3    (A7) I agree, performing a hemi-laryngectomy would give a cure rate as good as radiotherapy.
RT1    (A8) Yes, I have performed many hemi-laryngectomies, and when I reviewed my case load, the cure rate was
      97%, which is as good as that reported internationally for radiotherapy.
RT2   (A9) I agree, however, you fail to take into account the patient's age. Given the patient is over 75, operating on
      the patient is not advisable as the patient may not recover from an operation.
RT1   (A10) Yes, however, in this case, the patient's performance status is extremely good, the patient will most likely
      recover from an operation. (i.e. the general rule does not apply)
S2    (A11) Reviewing our past case decisions, evidence suggest that the we have always performed a hemi-
      laryngectomy, hence my preference is to do the same.
S3    (A12) I agree, however, there is some new medical literature reporting that the voice quality after a hemi-
      laryngectomy was only 50% acceptable and the reporting institution was the North American leaders in hemi-
      laryngectomy, hence we should perform radiotherapy.
```

Figure 6-2 Consultation Record of Larynx Therapy [73]

Assume this argumentation record is written in form of definition 6.3:

S1/A1    cure rate (99%) ← take out;

S2/A2    cure rate (99%)← take out;

S3/A3    cure rate (99%)← take out;

RT1/A4    Lose voice← take out;

RT1/A5    cure rate (97%), voice quality (97%) ← Radiotherapy;

S2/A6    cure rate (97%)← Hemi-laryngectomy;

S3/A7    cure rate (97%)← Hemi-laryngectomy;

RT1/A8    cure rate (97%), reported internationally ← Hemi-laryngectomy;

RT2/A9    will not recover (patient (over (age, 75)))← Hemi-laryngectomy,;

RT1/A10    will mostly recover (this patient) ← Hemi-laryngectomy;

S2/A11    personal experience ← Hemi-laryngectomy;

S3/A12    voice quality (50%) ←Hemi-laryngectomy.

This time we take the technical impact in account for elements comparison. Each element is only compared with other elements of the same sort. Summarize the above arguments, we have the following three sets:

Take Out: {Cure rate 99%, Voice quality 0%},
Radiotherapy: {Cure rate 97%, Voice quality 97%},
Hemi-laryngectomy :{Cure rate 97%, Voice quality 50%}.

The result partial order is shown in Figure 6-3 (a). This result may not satisfy the expectations of doctors and patients. A better model is to differentiate preferred elements from general ones. This can be fixed in the model or selected by doctors or patients. If we select cure rate as preferred, then the result is shown in Figure 6-3 (b), where the arrow means equal to or less than.

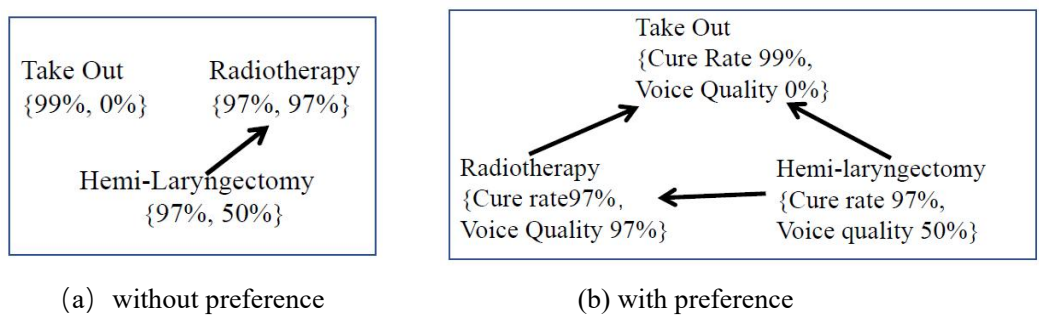

(a) without preference  (b) with preference

Figure 6-3 Partial orders without and with preference

## $7  Conclusion and further Remarks

### 7.1  A short summary

The main contribution of the paper can be summarized as follows.

1. We propose the concept of logic programming over knowledge graphs, and as a first attempt, we design and implement a Prolog style logic programming language KGN-Prolog, where KGN stands for Knowledge Graph Network for multiple KGs interconnected into a network structure. KGN-Prolog features logic programming on both triplets-structured systems and low dimensional vector spaces, unsharp and uncertain events, concurrent and distributed implementations, deep and multi-modal computation, consultation and argumentation interactions, free style and federated paradigms, and so on. All represented in KGN-Prolog.

2. We demonstrate the power of KGN-Prolog in a series of medical applications, including medical knowledge's representation and searching, management and completion, classification and learning, image processing and analysis, etc. Among them, medical consultation and argumentation is implemented through two-level partially ordered programming, which regards argumentation as preference competition. Real-world medical examples are used to illustrate these functions.

3. In order to implement the above functions, we have introduced new functions in logic programming. For connecting representations of logic knowledge with knowledge graphs, we introduce representation of triplets as comparative predicates. At the same time, we introduce the principle of partial unification and comparative predicates to bridge the gap between knowledge graph and logic rule representation. In order to overcome the enumeration complexity brought by large data sets, KGN-Prolog allows the fail operator be parametrized with a number or variable. It also allows variable value sequences listed in data part to simplify rule representation.

**7.2    KGNF-A multi-layered model for federated learning and reasoning on KGN**

In the big data (BD) era, data sharing and data protection are equally important. Also, data cooperation and data privacy have the same importance. Only paying attention to one of the two principles is not enough and even dangerous. The KGN framework introduced in the previous sections of this paper concentrates on KG's knowledge sharing and function cooperation aspects. But this is not enough. It is of significant importance to combine knowledge sharing with knowledge protection and to introduce conditioned cooperation instead of providing unconditional service. Many works have been done in this area. The most referenced work in this direction is federated machine learning proposed by [74], Recent publications include federated KG completion [75], federated KG embedding [76] and federated KG integration [77]. In this subsection we propose a framework of federated KGN, where the federation is designed on different levels. The general idea of KGN is to promote the former. But this idea is not perfect. It emphasizes only the cooperation aspect. In this sub-section we will introduce briefly the KGNF framework, which adds a set of additional specifications to KGN relating to the data protection aspect.

**Definition 7.1:** The KGNF framework is KGN plus a selective set of the following additional specifications:

1). Knowledge protection at individual entity level: person name, location, sensitive data, etc.

2). Knowledge Protection at group level: only part of the data in some area are free accessible. No statistics of data is allowed to be done. This group level can be a schema, a class, an area, or other data structures.

3). Knowledge protection at KG level: A KG can be visited only through its LPP program. In this way the safety of knowledge and data in the KG is protected by the LPP program. It not only decides which external KGs have the right of visiting its partner KG, but also decides which parts of the KG are allowed to be visited.

4). Knowledge protection at function level: For example the downloading of a KG or its data can be excluded by function protection.

**Example 7.1.** KGNF allows the rules from Example 2.2 (Section 2) to be rewritten as follows, where the KG limits the Generate1(X) rule's loop to at most 120 repetitions and the Generate2(X) rule to at most 130 repetitions. That means the snapshot's size is limited to at most 250 triples.

Snapshot (X) ← Generate1 (X); Genarate2 (X).
Generate1(X) ← Output ((RDF (X, Y, Z))), Fail (120).                         (7-1)
Generate2(X) ← Output ((RDF (U, V, X)), Fail (130).

**Example 7.2.** KGNF allows the rules of example 3.1 of section 3 rewritten in following form, where the vector-formed KG-V can only be visited through its partner LPP which checks the legality of visitor's requirement.

Snapshot(X) ← in-class (X, person), #KG-V-LPP# outputV (Vec (X, Y1, Z1)), Fail;
in-class    (X,    person),    #KG-V-LPP#    outputV    (Vec    (X2,    Y2,    X)),    Fail. (7-2)

where the KG-V-LPP rules are the following, which only allow civil people to be snapshotted:

outputV (Vec (X, Y1, Z1)) ← in-class (X, civil), outputV-LPP (Vec (X, Y1, Z1)), Fail;
outputV (Vec (X2, Y2, X)) ← in-class (X, civil), outputV-LPP (Vec (X2, Y2, X)), Fail.          (7-3)

Acknowledgement:    This research was supported in part by NSFC Project 61621003.

Literature

[1] K Murphy, From big data to big knowledge, 22nd ACM Int. Conf. Inf. Knowl. Manage., pp. 1917–1918, 2013.

[2] T Goodwin, S M Harabagiu, Automatic generation of a qualified medical knowledge graph and its usage for retrieving patient cohorts from electronic medical records, IEEE Seventh International Conference on Semantic Computing. pp. 363–370. IEEE (2013).

[3] P Ernst, C Meng, A Siu, G Weikum, Knowlife: a knowledge graph for health and life sciences. In: 2014 IEEE 30th International Conference on Data Engineering. IEEE (2014).

[4] L Jia, J Liu, Y Dong et al., Construction of traditional Chinese medicine knowledge graph, Journal of Medical Informatics, Vol 12, pp.51-59, 2015.

[5] L Li et al., Real-world data medical knowledge graph: construction and applications, Artificial Intelligence in Medicine, 2020.

[6] M Rotmensch et al., Learning a health knowledge graph from electronic medical records. Scientific reports 7(1), 5994 , 2017.

[7] B Cheng , J Zhang , H Liu et al., Research on Medical Knowledge Graph for Stroke, Journal of Healthcare Engineering,Vol. 2021, Article ID 5531327.

[8] A Harnoune, M Rhanoui, M Mikram et al., BERT based clinical knowledge extraction for biomedical knowledge graph construction and analysis, computer based methods and programs in biomedicine update, Vol 1, 100042, 2021.

[9] P Chandak, K Huang, M Zitnik, Building a knowledge graph to enable precision medicine, https://doi.org/10.1038/s41597-023-01960-3, 2023.

[10] Y Fan, Z Li. Research and Application Progress of Chinese Medical Knowledge Graph, Journal of Frontiers of Computer Science and Technology, 16(10): 2219-2233, 2022.

[11] S Zhang, R Lu et al., Patients – centered holographic medical and healthcare assistance key techniques and applications, 2020.

[12] F Daniel, B Volha, B Christian, DBpedia latest core released, https://databus.dbpedia.org/dbpedia/collections/latest-core, 2019.

[13] F Mahdisoltani, J Biega, F Suchanek. Yago3: A knowledge base from multilingual wikipedias[C]//7th biennial conference on innovative data systems research. CIDR Conference, 2014.

[14] E L Schiffrin, Beyond blood pressure: the endothelium and atherosclerosis progression[J]. American journal of hypertension, 2002, 15(S5): 115S-122S.

[15] M E Safar, P Jankowski, Central blood pressure and hypertension: role in cardiovascular risk assessment[J]. Clinical science, 2009, 116(4): 273-282.

[16] W Hollander, Role of hypertension in atherosclerosis and cardiovascular disease[J]. The American journal of cardiology, 1976, 38(6): 786-800.

[17] Cardiovascular Disease dataset. https://www.kaggle.com/datasets/sulianova/cardiovascular-disease-dataset.

[18] Centers for Disease Control and Prevention. About Adult BMI, https://www.cdc.gov/healthyweight/assessing/bmi/adult_bmi/index.html.

[19] X Han, S Cao, X Lv, et al. Openke: An open toolkit for knowledge embedding, Proc. of the conference on empirical methods in natural language processing: system demonstrations, pp. 139-144, 2018.

[20] S Dreiseitl, L O-Machado, Logistic regression and artificial neural network classification models: a methodology review, Journal of biomedical informatics, 35(5-6): pp.352-359, 2002.

[21] F Pedregosa, G Varoquaux, A Gramfort, et al. Scikit-learn: Machine learning in Python, the Journal of machine Learning research, V 12, pp. 2825-2830, 2011.

[22] L A Zadeh, "Fuzzy sets," Information and Control, Vol.8, pp. 338-353, 1965.

[23] L A Zadeh, The concept of a linguistic variable and its application to approximate reasoning, Parts 1, 2, and 3, Information Sciences, Vol. 8, pp.199-249, pp.301-357, Vol. 9, pp. 43-80, 1975.

[24] L A Zadeh, "Fuzzy logic = computing with words," IEEE Transactions on Fuzzy Systems, vol. 2, pp. 103–111, 1996.

[25] M Ishizuka, N Kanai, Prolog-ELF incorporating fuzzy logic, New Generation Computing, Vol 3, pp.479-486, 1985.

[26] R C T Lee, Fuzzy logic and the resolution principle, J. ACM, Vol. 19, pp.109-119, 1972.

[27] M Mukaidono, Z Shen, L Ding, Fundamentals of fuzzy Prolog, International J. of approximate reasoning, Vol. 3, pp.179-193, 1989.

[28] T P Martin, J F Baldwin, B W Pilsworth, The implementation of FProlog—A fuzzy Prolog interpreter, Fuzzy Sets and Systems, Vol. 23, pp. 119-129, 1987.

[29] S Guadarrama, S Muñoz, C Vaucheret, Fuzzy Prolog: a new approach using soft constraints propagation, Fuzzy Sets and Systems 144, pp.127–150, 2004.

[30] P Julián, C Rubio and J Gallardo. Bousi~Prolog: a Prolog extension language for flexible query answering. ENTCS, Vol 248, pp. 131-147, 2009.

[31] P Julián, C Rubio, An Efficient Fuzzy Unification Method and its Implementation into the Bousi~Prolog System, Proc. of FUZZ-IEEE, 2010.

[32] P J-Iranzo, F S-Perez, WordNet and Prolog: why not?, EUSFLAT pp. 827-834, 2019.

[33] P J Iranzo, G Moreno, J Penabad, Thresholded semantic framework for a fully integrated fuzzy logic language, J. Log. Algebraic Methods Program. 93, pp.42–67, 2020.

[34] P J Iranzo, G Moreno, J A Riaza, The fuzzy logic programming language FASILL: design and implementation, International Journal of Approximate Reasoning, 125, pp.139-168, 2017.

[35] V P-Ceruelo, S M-Hernandez, H Strass, Rfuzzy framework, WLPE2008, CoRR, abs/0903.2188, 2009.

[36] S M-Hernandez, V P-Ceruelo, H Strass, RFuzzy: Syntax, semantics and implementation details of a simple and expressive fuzzy tool over Prolog, Information Sciences 181, 1951–1970, 2011.

[37] J Medina, M O-Aciego, J R-Calvino, Fuzzy logic programming via multilattices, Fuzzy Sets and Systems 158, pp. 674 – 688, 2007.

[38] R Ng, V S Subrahmanian, Probabilistic logic programming, Information and Computation, 101, pp.150-201, 1992.

[39] T Lukasiewicz, Probabilistic logic programming with conditional constraints, ACM Transactions on Computational Logic, Vol. 2, No. 3, pp. 289–339, 2001.

[40] L D Raedt, A Kimmig, H Toivonen, ProbLog: A probabilistic Prolog and its application in link discovery, IJCAI-07, pp.2468-2473, 2007.

[41] A Dries, A Kimmig, W Meert et al., ProbLog2: Probabilistic logic programming, LNAI 9286, pp. 312–315, 2015.

[42] T Sato, Generative modeling by PRISM, in P M Hill, D S Warren (Eds.), Logic Programming, LNCS 5649, pp.24-35, 2009.

[43] J F Baldwin, Automated fuzzy and probabilistic inference, Fuzzy sets and systems, 18, pp.219-235., 1986.

[44] Internal technical report, 2021.

[45] M D Kohli, R M Summers, J R Geis, Medical Image Data and Datasets in the Era of Machine Learning—Whitepaper from the 2016 C-MIMI Meeting Dataset Session. J Digit Imaging. DOI 10.1007/s10278-017-9976-3, 2017.

[46] L O-Rayner, Exploring large scale public medical image datasets, http://arxiv.org/abs/1907.12720v1,2019.

[47] L O-Rayner, Exploring the ChestXray14 dataset: problems. 2018.

[48] P Rajpurkar, J Irvin, A Bagul, et al. Mura dataset: Towards radiologist-level abnormality detection in musculoskeletal radiographs, arXiv:171206957 2017.

[49] P R Cohen, Natural language techniques for multimodal interaction, Electronic Information Communication Society, Vol.J77-D2, No.8, pp.1403-1416, 1994.

[50] H Shimazu, Y Takashima, Multimodal definite clause grammar, Systems and Computers in Japan, 26 (3), pp. 93-102,1995.

[51] D B Moran, A J Cheyer, L E Julia et al., Multimodal User Interfaces in the Open Agent Architecture

[52] B G Batchelor, Merging the AUTOVIEW image processing language with PROLOG，Image and Vision Computing, 4 (4), pp. 189-196, 1986.

[53] Y Breux, S Druon, Multimodal Object-Based Environment Representation for Assistive Robotics, International Journal of Social Robotics, https://doi.org/10.1007/s12369-019-00600-4, 2019.

[54] S M Anwar, M Majid, A Qayyum, etc., Medical Image Analysis using Convolutional Neural Networks: A Review, arXiv:1709.02250v2 [cs.CV] 21, 2019.

[55] M Berman, H J´egou, A Vedaldi, I Kokkinos, M Douze, MultiGrain: a unified image embedding for classes and instances, arXiv:1902.05509v2 [cs.CV], 2019.

[56] K He, X Zhang, S Ren et al. Deep residual learning for image recognition[C]//Proceedings of the IEEE conference on computer vision and pattern recognition. 2016: 770-778.

[57] H Y Zhou, S Yu, C Bian et al. Comparing to learn: Surpassing imagenet pretraining on radiographs by comparing image representations[C]//Medical Image Computing and Computer Assisted Intervention–MICCAI 2020: 23rd International Conference, Part I 23. pp. 398-407, 2020.

[58] X Fan, R Craven, R Singer, F Toni, M Williams, Assumption-Based Argumentation for

Decision-Making with Preferences: A Medical Case Study, Imperial College London, University College Hospital, UK, 2013.

[59] R Craven, F Toni, C Cadar et al., Efficient Argumentation for Medical Decision-Making, International Conference on Principles of Knowledge Representation and Reasoning, pp.598-602, 2012.

[60] M A Grando, D Glasspool, A Boxwala, Argumentation logic for the flexible enactment of goal-based medical guidelines, Journal of Biomedical Informatics 45 (2012) pp. 938–949.

[61] M A Qassas, D Fogli, M Giacomin, G Guida, Analysis of Clinical Discussions Based on Argumentation Schemes, Conference on ENTERprise Information Systems, 2015.

[62] L Moss, Explaining anomalies: an approach to anomaly-driven revision of a theory. PhD thesis, University of Aberdeen, UK; 2010.

[63] M A Grando, L Moss, D Glasspool et al. Argumentation-logic for explaining anomalous patient responses to treatments, Proc. Artificial intelligence in medicine, vol. 6747, 2011. pp. 35–44.

[64] M A Grando, L Moss, D Sleeman, J Kinsella, Argumentation-logic for creating and explaining medical hypotheses, Artificial Intelligence in Medicine 58 (2013) pp.1– 13.

[65] D W Glasspool, J Fox, A Oettinger, J S-Spark, Argumentation in decision support for medical care planning for patients and clinicians, Advanced Computation Laboratory Large Cancer Research UK, 2005.

[66] J Si, H H Kong, S H Lee, Developing clinical reasoning skills through argumentation with the concept map method in medical problem-based learning, Interdisciplinary Journal of Problem-Based Learning, Volume 13Issue 1, 2019.

[67] J Lu, S P Lajoie, Supporting medical decision making with argumentation tools, Contemp. Educ. Psychol. 33 (2008) 425–442.

[68] C Tarrant, T Stokes, A M Colman, Models of the medical consultation: opportunities and limitations of a game theory perspective, Education and Training, 2004.

[69] P Dung, R Kowalski, F Toni, Dialectic proof procedures for assumption-based, admissible argumentation. AIJ 170, pp. 114–159, 2006.

[70] X Fan, R Craven, R Singer et al., Assumption-Based Argumentation for Decision-Making with Preferences: A Medical Case Study, pp. 374-390, Computational Logic in Multi-Agent Systems, LNAI 8143, pp.374-390, 2013.

[71] N Kökciyan, I Sassoon, E Sklar et al., Applying metalevel argumentation frameworks to support medical decision making, IEEE Intelligent Systems, DOI10.1109/MIS.2021.3051420, 2021.

[72] L Xiao, Towards evidence-based argumentation graph for clinical decision support, IEEE 35th International Symposium on Computer-Based Medical Systems (CBMS), pp. 400-405, doi: 10.1109/CBMS55023.2022.00078, 2022.

[73] M A Qassas, D Fogli, M Giacomin et al., Analysis of Clinical Discussions Based on Argumentation Schemes, Procedia Computer Science 64. pp. 282 – 289, 2015.

[74] Q Yang, Y Liu, T Chen, et al. Federated machine learning: Concept and applications, http://arxiv.org/abs/1902.04885, 2019.

[75] M Chen, W Zhang, Z Yuan, Y Jia, H Chen, Federated knowledge graph completion via embedding-contrastive learning, Knowledge Based Systems, Vol 252, 109459, 2022.

[76] H Peng, H Li, Y Song, V Zheng, J Li, Differentially Private Federated Knowledge Graphs

Embedding, arXiv:2105.07615v2 [cs.LG], 2021.

[77] C Bizon, S Cox, J Balhoff, et al., ROBOKOP KG and KGB: Integrated Knowledge Graphs from Federated Sources, J. Chem. Inf. Model. 59, 4968−4973, 2019

[78] X-F Niu, W-J Li, ParaGraph E: A Library for Parallel Knowledge Graph Embedding, arXiv:1703.05614v3 [cs.AI] 5 Apr 2017.

[79] D Zhang, M Li, Y Jia, Y Wang, X Cheng, Efficient parallel translating embedding for knowledge graphs, arXiv:1703.10316v4 [cs.AI], 2018.

[80] C J Hsieh, H F Yu, I Dhillon. Passcode: Parallel asynchronous stochastic dual co-ordinate descent[C]//International Conference on Machine Learning. PMLR, 2015: 2370-2379.

[81] **B** Recht, C Re, S Wright et al. Hogwild!: A lock-free approach to parallelizing stochastic gradient descent. Advances in neural information processing systems, 2011, 24. pp. 693-701.